\PassOptionsToPackage{most,breakable}{tcolorbox}
\PassOptionsToPackage{dvipsnames,table}{xcolor}
\PassOptionsToPackage{numbers,compress}{natbib}
\documentclass[11pt]{article}

\usepackage[final]{ropedia}

\usepackage[T1]{fontenc}
\usepackage[utf8]{inputenc}
\usepackage{amsmath,amssymb}
\usepackage{amsfonts}
\usepackage{mathpazo}
\usepackage[xcharter,bigdelims,vvarbb]{newtxmath}
\usepackage[scaled]{helvet}
\usepackage[scaled=1.1]{zlmtt}
\usepackage{latexsym}
\usepackage{pifont}
\usepackage{placeins}
\usepackage{booktabs}
\usepackage{multirow}
\usepackage{tabularx}
\usepackage{subcaption}
\usepackage{adjustbox}
\usepackage{makecell}
\usepackage{enumitem}
\usepackage{rotating}
\usepackage{titletoc}
\usepackage{comment}
\usepackage{wrapfig}
\usepackage{needspace}
\usepackage{nicefrac}
\usepackage{microtype}
\usepackage{siunitx}
\usepackage{dblfloatfix}
\usepackage{graphicx}
\usepackage{cuted}
\usepackage{capt-of}
\usepackage{xspace}
\setlist[itemize]{leftmargin=2em}

\usepackage{tikz}
\usetikzlibrary{fadings}
\usetikzlibrary{decorations.text}
\usepackage{array}
\usepackage{tcolorbox}
\usepackage{threeparttable}

\newcommand{\projecttitle}{EgoTools: Towards Tool-Centric Reasoning\xspace}
\newcommand{\projectsubtitle}{in Real-World Egocentric Videos\xspace}

\newcommand{\ourmethod}{EgoTools\xspace}
\newcommand{\ourdata}{EgoTools-Data\xspace}
\newcommand{\ourbench}{EgoTools-Bench\xspace}
\newcommand{\ourmodel}{\ourmethod}

\newcommand{\cmark}{\textcolor{green!60!black}{\ding{51}}}
\newcommand{\xmark}{\textcolor{red!70!black}{\ding{55}}}

\definecolor{subcol}{HTML}{F7FFF2}

\colorlet{rankfirst}{ropediagreen}
\colorlet{ranksecond}{ropedialightgreen}
\colorlet{rankthird}{ropediaabstractgreen}

\colorlet{bestgreen}{rankfirst}
\colorlet{secondgreen}{ranksecond}
\colorlet{thirdgreen}{rankthird}

\colorlet{bestred}{rankfirst}
\colorlet{secondorange}{ranksecond}
\colorlet{thirdyellow}{rankthird}

\definecolor{trainmark}{RGB}{0,180,0}
\definecolor{oomred}{RGB}{252,228,228}
\definecolor{catgray}{HTML}{E7F8DC}
\definecolor{altcolgreen}{HTML}{F0FFE8}
\colorlet{altcolblue}{altcolgreen}
\definecolor{pretrainmark}{RGB}{197,90,17}
\definecolor{gaingreen}{RGB}{30,140,60}
\definecolor{lossred}{RGB}{200,30,30}
\definecolor{navyblue}{HTML}{4F8F35}

\colorlet{findingback}{ropedialightgreen}
\colorlet{findingbacksoft}{ropediaabstractgreen}
\colorlet{findingframe}{ropediadeepgreen}
\colorlet{findingaccent}{ropediaaccent}

\newcommand{\appendixtablestyle}{\small
  \setlength{\tabcolsep}{5pt}\renewcommand{\arraystretch}{1.10}\captionsetup{skip=5pt}}

\definecolor{darkblue}{RGB}{0, 0, 139}

\newcommand{\pmark}{\raisebox{-0.14em}{\includegraphics[height=1.05em]{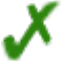}}}

\DeclareUnicodeCharacter{237B}{\pmark}

\newcommand{\sigIcon}[1]{\raisebox{-0.14em}{#1}}
\newcommand{\sigVideo}{\sigIcon{\includegraphics[height=1.05em]{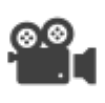}}}
\newcommand{\sigThree}{\sigIcon{\includegraphics[height=1.05em]{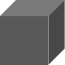}}}
\newcommand{\sigImu}{\sigIcon{\includegraphics[height=1.05em]{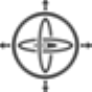}}}
\newcommand{\sigGaze}{\sigIcon{\includegraphics[height=1.05em]{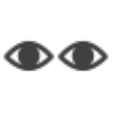}}}

\definecolor{PerfGreen}{HTML}{95D5B2}
\newcommand{\best}[1]{\cellcolor{PerfGreen!55}\textbf{#1}}
\newcommand{\second}[1]{\cellcolor{PerfGreen!20}#1}
\newcommand{\third}[1]{\cellcolor{PerfGreen!10}#1}

\newcommand{\modelgroup}[1]{\rowcolor{gray!20} \multicolumn{11}{c}{\textbf{\textit{#1}}}
}

\definecolor{DeltaGreen}{HTML}{2E7D32}
\definecolor{DeltaRed}{HTML}{B23A48}

\newcommand{\gain}[1]{\textcolor{DeltaGreen}{\textbf{#1}}}
\newcommand{\drop}[1]{\textcolor{DeltaRed}{\textbf{#1}}}

\firstpagelogos{\includegraphics[height=15pt]{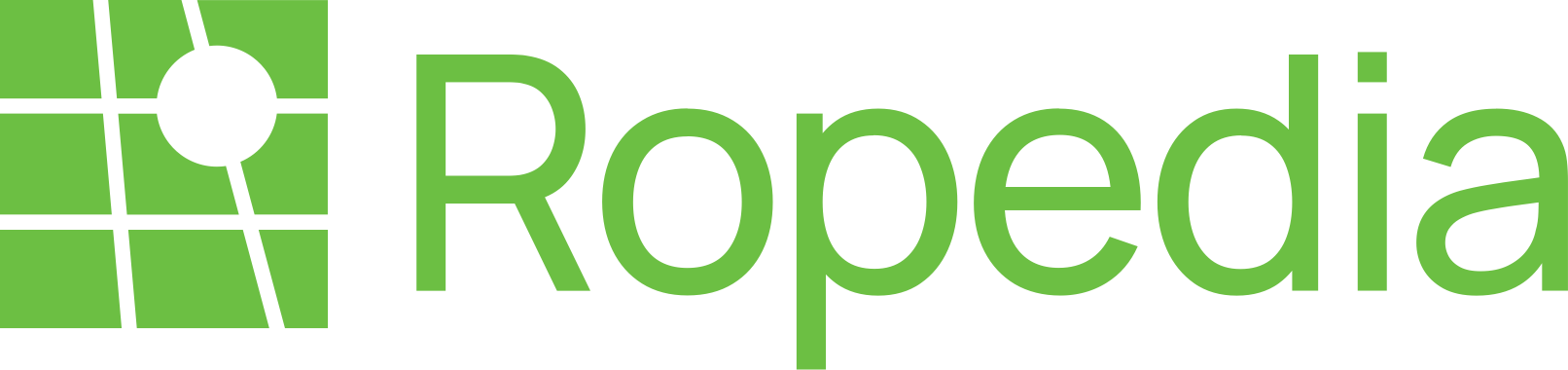}}{\includegraphics[height=15pt]{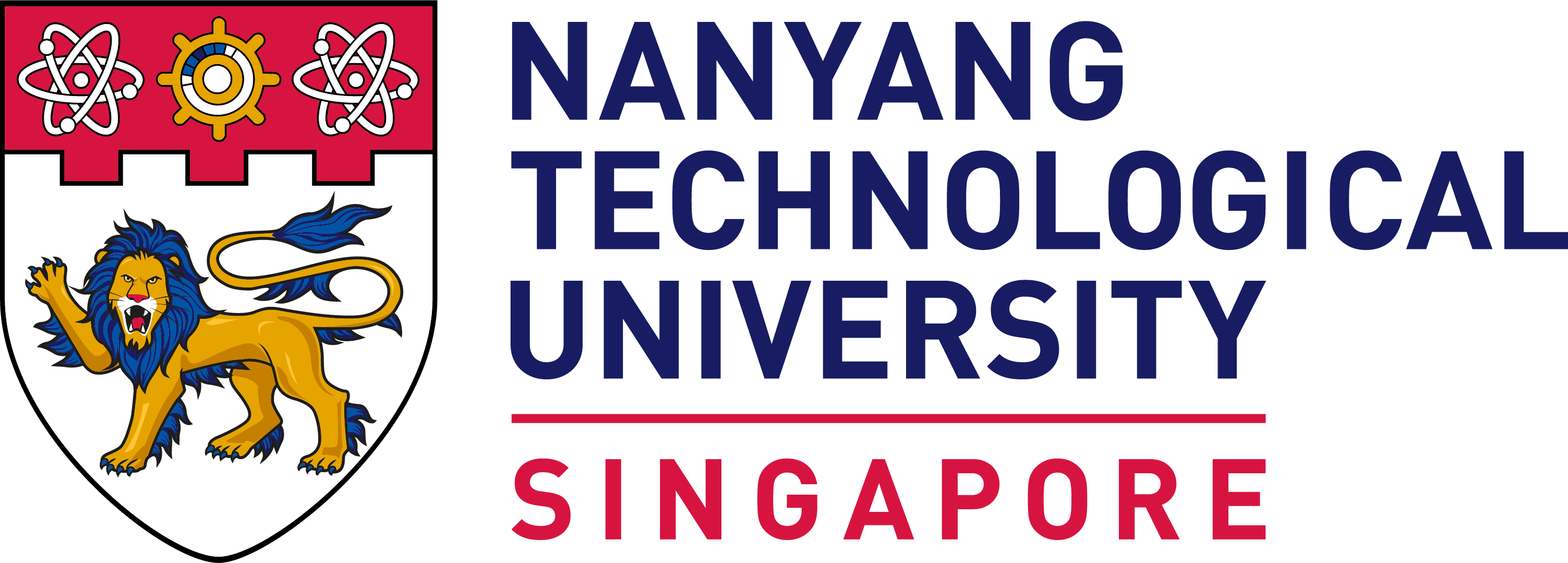}}{}
\abstractlogo{\includegraphics[height=18pt]{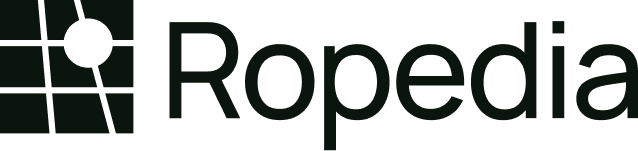}}

\title{\projecttitle}
\subtitle{\projectsubtitle}

\newcommand{\ropediaauthormark}{\textsuperscript{\raisebox{0.05ex}{\scalebox{1.4}{\textcolor{ropediaaccent}{$\star$}}}}}
\newcommand{\authorfootnotes}{\begingroup
  \renewcommand{\thefootnote}{}\footnotetext{* Equal contribution. $\dagger$ Corresponding author. \ropediaauthormark{} Ropedia Author.}\endgroup
  \setcounter{footnote}{0}}

\author{
  \AuthorName{Shulin Tian}{1,2,*}\quad
  \AuthorName{Junsu Kim}{1,3,*}\quad
  \AuthorName{Shuai Liu}{1,*}\quad
  \RopediaAuthorName{Hao Li}{1,*}\quad
  \AuthorName{Yujiao Shen}{1}\quad
  \AuthorName{Sihan Li}{1}\quad
  \AuthorName{Zhe Yang}{1}\quad
  \AuthorName{Yeongon Kim}{1}\quad
  \AuthorName{Feiyu Li}{4}\quad
  \AuthorName{Jialin Wu}{5}\quad
  \AuthorName{Yichi Zhang}{1}\quad
  \AuthorName{Wenhui Wang}{6}\quad
  \AuthorName{Runmao Yao}{1}\quad
  \AuthorName{Yuhao Dong}{1}\quad
  \RopediaAuthorName{Zhaoxi Chen}{1}\quad
  \RopediaAuthorName{Fangzhou Hong}{1}\quad
  \AuthorName{Antonino Furnari}{7}\quad
  \AuthorName{Jingkang Yang}{1}\quad
  \AuthorName{Hongyuan Zhu}{2}\quad
  \RopediaAuthorName{Ziwei Liu}{1,$\dagger$}\\[1.0ex]
  {\normalfont\footnotesize
  \mbox{\textsuperscript{1}S-Lab, Nanyang Technological University}\quad
  \mbox{\textsuperscript{2}A*STAR}\quad
  \mbox{\textsuperscript{3}KAIST}\quad
  \mbox{\textsuperscript{4}PKU}\quad
  \mbox{\textsuperscript{5}FDU}\quad
  \mbox{\textsuperscript{6}School of Biological Sciences, Nanyang Technological University}\quad
  \mbox{\textsuperscript{7}University of Catania}
}}

\newcommand{\teaserplaceholder}{\begin{tcolorbox}[
      colback=ropediaabstractgreen,
      colframe=ropediaaccent,
      coltext=ropediaink,
      boxrule=0.6pt,
      width=\textwidth,
      height=0.32\textwidth,
      valign=center,
      arc=2mm,
      left=6mm,
      right=6mm
    ]
    \centering\sffamily
    {\Large\bfseries Teaser Placeholder}\\[0.6ex]
    Replace this box by adding \texttt{figs/teaser.pdf}.
  \end{tcolorbox}}

\begin{document}
\maketitle
\authorfootnotes

\begin{paperresources}
\raggedright
\paperresourceicon{\makebox[1em][c]{\resourceprojecticon}}{Project Page}{https://ropedia.github.io/egotools}{ropedia.github.io/egotools}\\[2pt]
\paperresourceicon{\makebox[1em][c]{\resourcegithubicon}}{Code}{https://github.com/Ropedia/EgoTools}{github.com/Ropedia/EgoTools}\\[2pt]
\paperresourceicon{\makebox[1em][c]{\resourcehficon}}{Dataset}{https://huggingface.co/ropedia-ai/egotools-data}{huggingface.co/ropedia-ai/egotools-data}\\[2pt]
\paperresourceicon{\makebox[1em][c]{\resourcehficon}}{Model}{https://huggingface.co/ropedia-ai/egotools-8b}{huggingface.co/ropedia-ai/egotools-8b}
\end{paperresources}

\vspace{6pt}
{\centering
\IfFileExists{fig/teaser_egotools.pdf}{\includegraphics[width=1\textwidth]{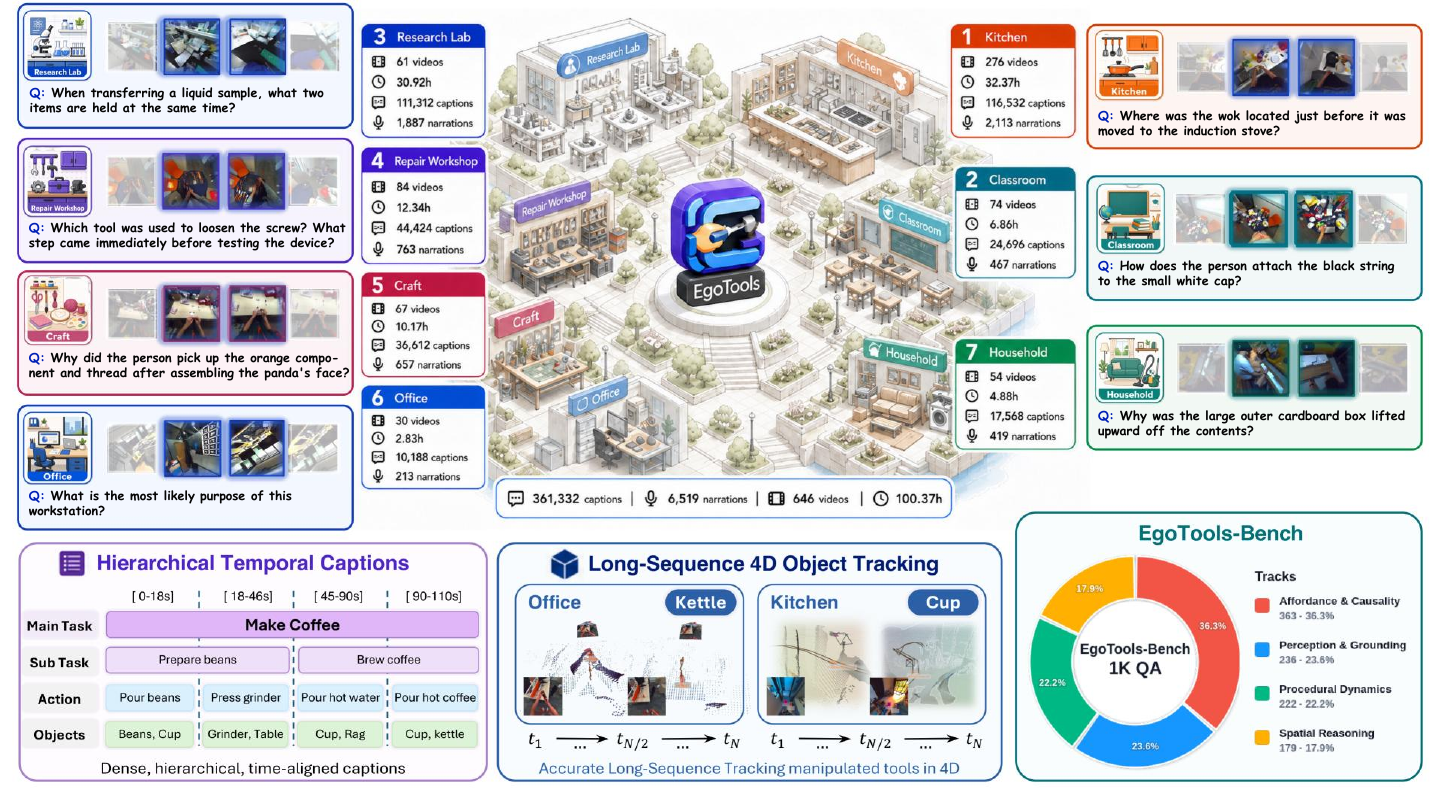}}{\teaserplaceholder
}
\par\vspace{5pt}
\begin{minipage}{\textwidth}
\footnotesize
\refstepcounter{figure}\label{fig:teaser}
Figure 1: \textbf{Overview of EgoTools.} EgoTools is an egocentric tool-use dataset that spans multiple environments, with dense hierarchical captions and long-sequence 4D object tracking. We also propose \textbf{EgoTools-Bench}, a 1,000-question benchmark spanning diverse real-world environments and reasoning tasks.
\end{minipage}
\par}
\vspace{6pt}

\Needspace{8\baselineskip}
\begin{abstract}
  Real-world embodied tasks, from everyday activities to professional procedures, require agents to act under physical constraints while tracking evolving object and task states. Tool use sits at the heart of such tasks, as many everyday and professional activities are tool-mediated. Understanding them requires reasoning about affordances, hand--tool--object geometry, procedural progress, and causal effects on target objects. Yet despite strong performance on perception-oriented video tasks such as captioning and general video QA, current multimodal video models remain limited in this form of tool-centric embodied reasoning.
  Progress in this direction has been limited by the lack of real-world egocentric data and diagnostic benchmarks.
  To address this gap, we introduce \textbf{\ourmethod{}}, the first comprehensive suite for egocentric tool-use understanding. It consists
  of two complementary components: \textbf{EgoTools-Data}, a large-scale corpus of 100 hours of tool-centric egocentric recordings with
  synchronized audio, dense captions, reasoning-heavy narrations, and supplementary 3D information; and \textbf{EgoTools-Bench}, a
  diagnostic benchmark of 1,000 QA pairs across four tracks that cover tool-use understanding from perception and geometry to procedure and
  causal reasoning.
  Experimental results show that current models still struggle to ground tool use in visual evidence: Gemini-3.1-Pro achieves 66.9\% overall accuracy but only 51.7\% on Perception \& Grounding. Beyond evaluation, we validate \textbf{EgoTools-Data} as a training resource. On the full 1,000-question benchmark, full supervised fine-tuning improves Qwen3-VL-8B-Instruct from \textbf{50.0\% to 60.9\%}, under strict source-video separation. Together, these results establish \ourmethod{} as a unified resource for both training and diagnostic evaluation of real-world egocentric tool-use understanding.
\end{abstract}

\section{Introduction}
\label{sec:intro}

Human activities in the physical world, from daily chores to professional procedures, involve not only bare-hand manipulation but also the skilled use of tools. For an embodied agent, understanding such activities demands more than simply recognizing visible actions and objects. Such understanding also requires models to reason about tool--target contact, spatial coordination, procedural progress, and the physical outcomes of actions. This tight coupling of low-level perception with high-level temporal and causal awareness makes tool use a distinct challenge for embodied reasoning. The egocentric viewpoint naturally foregrounds this challenge by centering the observations of hands, tools, and their immediate consequences in a continuously evolving task state. From this perspective, the interplay of physical form, functional intent, and procedural dynamics is directly and persistently visible, making egocentric tool use a uniquely demanding and revealing testbed for embodied reasoning.

Despite strong performance on perception-oriented video tasks, current multimodal video models~\cite{bai2025qwen3vltechnicalreport,bai2025qwen25vltechnicalreport,team2024gemini,li2024llavaonevisioneasyvisualtask,zhu2025internvl3exploringadvancedtraining} remain limited in egocentric tool-use reasoning. Progress in this important direction has been held back by the lack of real-world egocentric data with dense tool-use annotations and by the absence of diagnostic benchmarks that isolate this reasoning. Most existing egocentric datasets~\cite{damen2021rescaling,grauman2022ego4d,grauman2024egoexo4d,liu2022hoi4d, ragusa2023meccano, sener2022assembly101} focus on activities, objects, or hand--object interactions. Although tools are often present in these videos, they are rarely foregrounded as the central unit of explanation. In previous benchmarks~\cite{chen2024egoplanbench,cheng2024egothink,di2024groundvqa,jia2022egotaskqa,mangalam2023egoschema,plizzari2025egotempo}, actions may be labeled or queried at the level of ``cook eggs'', while the spatula and pan that mediate the activity go unmentioned, and the tools through which actions unfold are effectively invisible in their annotations. This dual absence creates both an evaluation gap and a supervision gap: current benchmarks do not isolate tool-mediated reasoning, and existing egocentric corpora rarely provide dense training signals that connect tool choice, target-object state, manipulation context, and causal outcomes.

To address this gap, we introduce \textbf{\ourmethod{}}, the first comprehensive suite that brings together dedicated resources for both training and evaluation of egocentric tool-use understanding. At its core, EgoTools comprises two complementary components: \textbf{EgoTools-Data}, a large-scale real-world egocentric corpus designed to provide the rich training signals that current models lack, and \textbf{EgoTools-Bench}, a carefully constructed diagnostic benchmark that isolates the specific reasoning demands of tool-mediated activities for rigorous evaluation. EgoTools‑Data comprises 100 hours of tool‑centric egocentric recordings spanning diverse everyday and professional tasks, with synchronized audio, dense textual annotations ranging from surface-level captions to narrations that explicitly foreground affordances, procedural structure, and causal effects, as well as supplementary 3D information.
Its annotations expose learnable signals for tool choice and substitution, object-state changes, procedural progress, and grounded hand--tool--object interactions. Together, these annotations transform passive video into structured supervision for understanding how humans conduct embodied tasks with tools in the physical world. From this corpus, we construct EgoTools‑Bench, a set of 1,000 question--answer pairs that systematically probe a model's capacity for embodied tool use. Our benchmark consists of four tracks: \textbf{Affordance \& Causality} evaluates goal-directed reasoning about tool use, including why a tool is chosen or switched, what it affords, and how actions establish preconditions and produce consequences; \textbf{Perception \& Grounding} focuses on directly observable facts such as tool and object identities, attributes, states, and quantities; \textbf{Procedural Dynamics} probes the observable organization of tool-use processes over time, including tool--action ordering, step transitions, workspace staging, and fine-grained manipulation; and \textbf{Spatial Reasoning} evaluates egocentric geometric understanding of hand--tool--object relations, including position, alignment, containment, support, contact, and depth. These tracks provide a comprehensive evaluation of tool-use understanding from perception and geometry to procedure and causal reasoning.

We evaluate representative multimodal video models on EgoTools-Bench. On the quality-controlled full 1,000-question benchmark, Qwen3-VL-8B-Instruct reaches 50.0\%, indicating that substantial room remains for egocentric tool-use reasoning. We then test whether the annotations in EgoTools-Data constitute actionable supervision rather than serving only as an intermediate source for benchmark construction. Under strict source-video separation, fine-tuning Qwen3-VL-8B-Instruct on instruction data derived from the non-benchmark portion of EgoTools-Data improves accuracy from 50.0\% to 60.9\%. The model improves on three of the four reasoning tracks, with Spatial Reasoning the exception. Thus, EgoTools not only exposes a substantial gap in current video-language models, but also provides supervision that can partially close this tool-centric reasoning gap.

\section{Related Work}
\label{sec:related}

\begin{table*}[!t]
\centering
\caption{
\textbf{Comparison with representative egocentric datasets and benchmarks.}
A checkmark indicates that the feature is a central design component; \pmark{} indicates partial support. Signals denote sensor/video inputs and exclude QA text or narrations used for benchmark construction: \sigVideo{} = RGB/video, \sigThree{} = 3D/depth/pose, \sigImu{} = inertial signals, and \sigGaze{} = gaze; non-RGB signals may be available only for subsets of large datasets. Hours are reported total video hours or computed from source-reported clip counts and durations; ``--'' denotes not applicable or not reported. Embodied interaction benchmarks are included as scope contrast rather than directly comparable egocentric video corpora.
}
\setlength{\tabcolsep}{3.0pt}
\renewcommand{\arraystretch}{1.12}
\begin{adjustbox}{max width=\textwidth}
\begin{tabular}{@{\hskip 5pt} l
!{\color{black!15}\vrule width 0.6pt}
l c c c
!{\color{black!15}\vrule width 0.6pt}
c c c
!{\color{black!15}\vrule width 0.6pt}
c c c
@{\hskip 5pt}}
\toprule
\makecell[l]{\textbf{Dataset / Benchmark}}
& \makecell[l]{\textbf{Scene}}
& \makecell{\textbf{Video}\\\textbf{Hours}}
& \makecell{\textbf{\#QA}}
& \makecell{\textbf{Signals}}
& \makecell{\textbf{Multi-}\\\textbf{Env.}}
& \makecell{\textbf{Tool-Centric}\\\textbf{Narration}}
& \makecell{\textbf{Narr.-Linked}\\\textbf{Tool Grounding}}
& \makecell{\textbf{QA}\\\textbf{Benchmark}}
& \makecell{\textbf{Tool Choice /}\\\textbf{Substitution}}
& \makecell{\textbf{Physical Tool-}\\\textbf{Use Reasoning}} \\
\midrule
\rowcolor{black!6}
\multicolumn{11}{@{\hskip 5pt}l@{\hskip 5pt}}{\textit{\small — Egocentric Observation Datasets —}} \\
\noalign{\vspace{2pt}}
EPIC-KITCHENS~\cite{damen2021rescaling}
& Kitchen & 100 & -- & \sigVideo
& \xmark & \xmark & \xmark & \xmark & \xmark & \xmark \\
Ego4D~\cite{grauman2022ego4d}
& Real-world & 3,670 & 14.5\,K & \sigVideo\,\sigThree\,\sigImu\,\sigGaze
& \cmark & \xmark & \xmark & \pmark & \xmark & \xmark \\
Ego-Exo4D~\cite{grauman2024egoexo4d}
& Skilled & 1,286 & -- & \sigVideo\,\sigThree\,\sigImu\,\sigGaze
& \cmark & \xmark & \xmark & \xmark & \xmark & \pmark \\
Assembly101~\cite{sener2022assembly101}
& Assembly & 513 & -- & \sigVideo\,\sigThree
& \xmark & \xmark & \xmark & \xmark & \xmark & \xmark \\
MECCANO~\cite{ragusa2023meccano}
& Industrial & 6.9 & -- & \sigVideo\,\sigThree\,\sigGaze
& \xmark & \xmark & \xmark & \xmark & \xmark & \pmark \\
HOI4D~\cite{liu2022hoi4d}
& Indoor HOI & 44.4 & -- & \sigVideo\,\sigThree
& \xmark & \xmark & \xmark & \xmark & \xmark & \pmark \\
\midrule
\rowcolor{black!6}
\multicolumn{11}{@{\hskip 5pt}l@{\hskip 5pt}}{\textit{\small — Egocentric Video Reasoning Benchmarks —}} \\
\noalign{\vspace{2pt}}
EgoSchema~\cite{mangalam2023egoschema}
& Real-world & 250+ & 5,031 & \sigVideo
& \cmark & \xmark & \xmark & \cmark & \xmark & \xmark \\
EgoTaskQA~\cite{jia2022egotaskqa}
& Task videos & 14 & 40\,K & \sigVideo
& \xmark & \xmark & \xmark & \cmark & \xmark & \pmark \\
EgoPlan-Bench~\cite{chen2024egoplanbench}
& Daily tasks & -- & 4,939 & \sigVideo
& \xmark & \xmark & \xmark & \cmark & \xmark & \pmark \\
EgoIntent~\cite{pan2026egointent}
& Daily tasks & 2.9 & 3,014 & \sigVideo
& \xmark & \xmark & \xmark & \cmark & \pmark & \pmark \\
GroundVQA~\cite{di2024groundvqa}
& Long videos & 736 & 303\,K & \sigVideo
& \xmark & \xmark & \xmark & \cmark & \xmark & \xmark \\
EgoTempo~\cite{plizzari2025egotempo}
& Temporal & 6.3 & 500 & \sigVideo
& \xmark & \xmark & \xmark & \cmark & \pmark & \pmark \\
\midrule
\rowcolor{black!6}
\multicolumn{11}{@{\hskip 5pt}l@{\hskip 5pt}}{\textit{\small — Embodied Interaction Benchmarks —}} \\
\noalign{\vspace{2pt}}
OpenEQA~\cite{majumdar2024openeqa}
& Indoor EQA & -- & 1.6\,K & \sigVideo\,\sigThree
& \cmark & \xmark & \xmark & \cmark & \xmark & \xmark \\
RoboVQA~\cite{sermanet2023robovqa}
& Robotics & 238 & 829\,K & \sigVideo
& \pmark & \xmark & \xmark & \cmark & \pmark & \pmark \\
RoboCasa365~\cite{nasiriany2026robocasa365}
& Sim kitchens & 2,200+ & -- & \sigVideo\,\sigThree
& \cmark & \xmark & \xmark & \xmark & \pmark & \pmark \\
\specialrule{1pt}{2pt}{0pt}
\textbf{EgoTools (Ours)}
& \textbf{Real-world} & 100 & 1,000 & \sigVideo\,\sigThree\,\sigImu
& \cmark & \cmark & \cmark & \cmark & \cmark & \cmark \\
\bottomrule
\end{tabular}
\end{adjustbox}
\label{tab:benchmark_comparison}
\vspace{-0.6em}
\end{table*}

\noindent\textbf{Egocentric video datasets and benchmarks.}
Egocentric video provides the actor's visual evidence, including hands, manipulated objects, surrounding context, and temporally ordered actions~\citep{bambach2015lending,fathi2012learning,li2018eye,pirsiavash2012detecting}. Large-scale datasets such as EPIC-KITCHENS, Ego4D, Ego-Exo4D, Assembly101, MECCANO, and HOI4D have enabled first-person action, object, hand-object, procedural, industrial, and skilled-activity understanding~\citep{damen2021rescaling,grauman2022ego4d,grauman2024egoexo4d,liu2022hoi4d,ragusa2023meccano,sener2022assembly101}. Recent video-language benchmarks further evaluate long-context QA, temporal grounding, planning, assistance, first-person reasoning, and long-form egocentric understanding~\citep{chen2024egoplanbench,cheng2024egothink,di2024groundvqa,jia2022egotaskqa,mangalam2023egoschema,plizzari2025egotempo,yang2025egolifeegocentriclifeassistant,zhou2025xlebench}. However, these resources are typically organized around activities, objects, events, or plans, leaving explicit tool-use reasoning under-specified. Closely related instructional-video tasks study detour retrieval and step differences involving ingredients, tools, or techniques~\citep{ashutosh2024detours,nagarajan2024stepdiff}, but they do not systematically test whether models can justify tool choice, compare feasible substitutes, or adapt manipulation to changing object states. \ourdata{} pairs egocentric recordings with participant-provided tool-centric narrations and focal-tool grounding. \ourbench{} evaluates complementary aspects of tool-use understanding from held-out egocentric videos.

\noindent\textbf{Physical tool-use reasoning.}
Physical tool use requires reasoning beyond object categories: a model must connect a tool's function and affordances to the current goal, target material, contact and motion constraints, and evolving object state. Robotics has studied related problems through tool-use surveys, task-oriented grasping, tool-flow prediction, affordance-centric manipulation, and egocentric affordance learning~\citep{fang2019taskoriented,li2025learningprecise,qin2023robottooluse,seita2022toolflownet,yamanobe2017brief}. These works are valuable for execution, but they often focus on structured tasks, constrained manipulation, or short interaction episodes. A parallel line evaluates MLLMs and VLMs on affordance grounding and physical tool understanding~\citep{huang2024manipvqa,qian2024affordancellm,yu2025seqafford,zhang2025phystoolbench}; many such settings rely on static images, 3D scenes, synthetic data, isolated interactions, or robotics setups. \ourbench{} complements both lines with human-centered first-person evidence, where real tools are selected, substituted, and adapted within continuous tasks. Accordingly, \ourmethod{} evaluates whether models can connect visual evidence to functional choices, feasible alternatives, temporal manipulation steps, and changes in object state.

\section{EgoTools Data Suite}
\label{sec:data}

\begin{figure*}[t!]
    \centering
    \includegraphics[width=1.\linewidth]{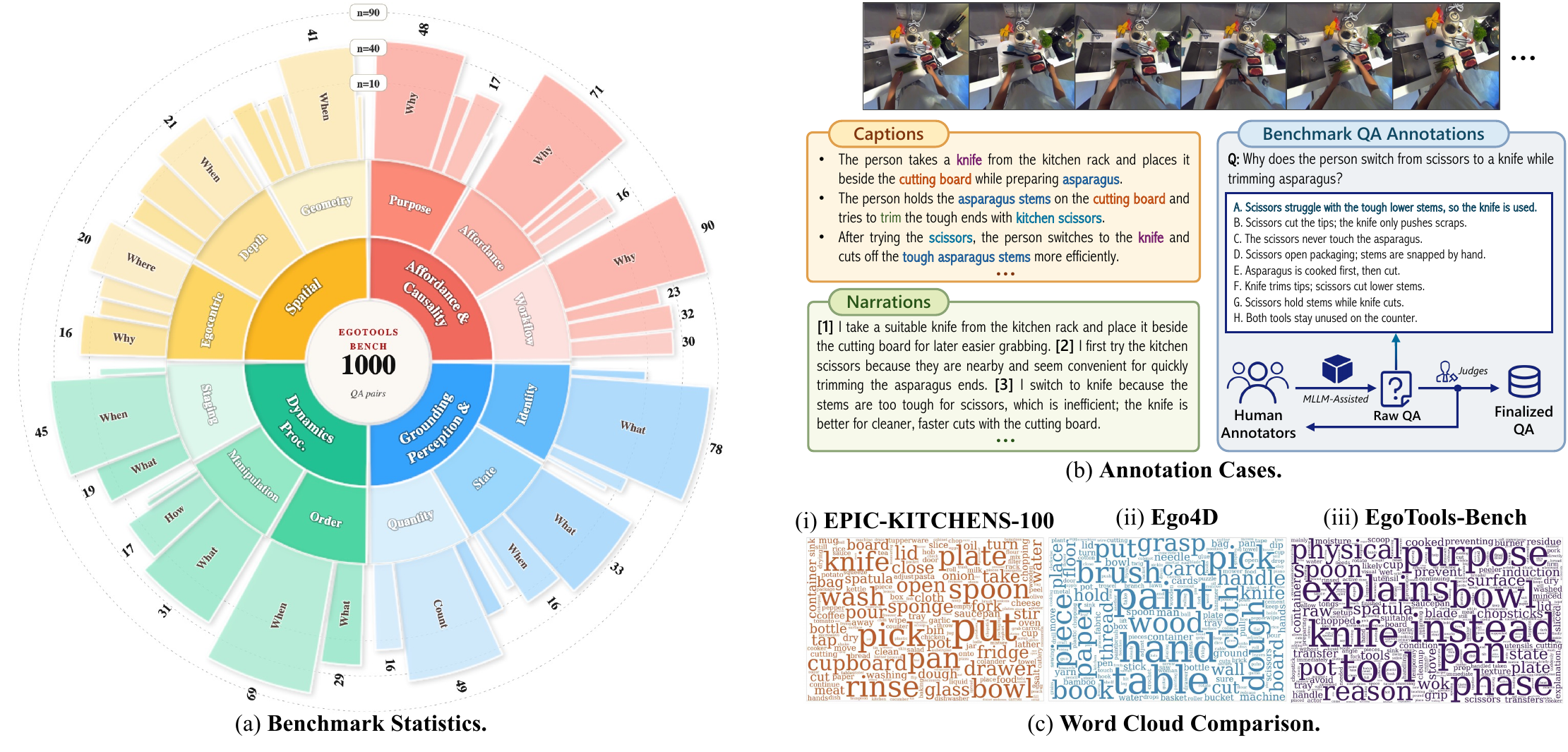}
    \caption{
    \textbf{Benchmark distribution and annotation examples.}
    (a) Distribution of 1,000 QA pairs across four tool-use reasoning tracks and their fine-grained subtracks.
    (b) Example annotations, including captions, tool-centric narrations, and finalized QA pairs produced through human annotation with MLLM-assisted checking.
    (c) Word-cloud comparison with EPIC-KITCHENS-100 and Ego4D, showing that \ourbench{} is more tool-dense and centered on tools, actions, objects, and state changes.
    }
    \label{fig:data_stat}
\end{figure*}

This section introduces \ourmethod{}, a real-world egocentric video suite for physically grounded tool-use understanding. We first describe the collection and preprocessing of \ourdata{}, a 100-hour real-world egocentric video corpus captured with synchronized video, audio, and geometry-related signals. We then introduce its multimodal annotation pipeline, including hierarchical dense captions, tool-centric narrations with 2D grounding, and 3D annotations derived from reconstruction and long-horizon object tracking. Finally, we describe two complementary resources derived from \ourdata{} and separated at the source-video level: (i) an EgoTools-derived video instruction-tuning corpus constructed from the training pool, and (ii) \ourbench{}, a 1,000-question 8-way multiple-choice diagnostic benchmark constructed from a curated 40.34-hour benchmark-reserved pool. \ourbench{} comprises 900 human-crafted questions and 100 human-verified spatial questions across four tracks: Affordance \& Causality, Perception \& Grounding, Procedural Dynamics, and Spatial Reasoning.

\subsection{Data Collection and Preprocessing}
\label{ssec:data_collection}

We collect raw egocentric videos with \textbf{HOMIE}\footnote{\url{https://ropedia.com/blog/20251216_introducing_ropedia}}, a lightweight head-mounted multimodal recording device designed with four synchronized fisheye camera views, together with audio and auxiliary motion/synchronization signals that support downstream spatial processing. These signals support 3D processing, including reconstruction, pose/depth estimation, and object/hand tracking. For annotation and training, we use the rectified front-left RGB stream as the canonical view, applying zoom and pitch adjustments to obtain a natural first-person perspective while preserving hand–tool–object interactions. Rectification details and examples are provided in Appendix~\ref{app:view_rectification}. The final videos are $1024 \times 1024$, 20 FPS, single-view egocentric streams synchronized with audio.

Our data collection is organized around two broad contexts: \textbf{daily activities} and \textbf{expertise-intensive procedures}. Within these contexts, we collect videos across seven tool-use domains: kitchen, classroom, research lab, repair workshop, craft, office, and household, shown in Figure~\ref{fig:teaser}.
These domains cover diverse tool-mediated activities, from everyday manipulation to craft, scientific, and fabrication procedures. Household recordings capture everyday tool use, craft and repair-workshop recordings emphasize material transformation and manual operations, and laboratory recordings include both educational and professional experiments requiring specialized knowledge and expert tool handling.
Data collection was conducted across multiple kitchens, workshops, laboratories, and daily-living spaces at universities and research sites in Asia. Our data collectors include graduate and undergraduate students from diverse disciplinary backgrounds, with domain expertise matched to the task whenever specialized knowledge is required. In particular, expertise-intensive recordings are performed or reviewed by collectors familiar with the corresponding procedures, ensuring that the captured tool use is both natural and technically valid. Details of the collection domains and anonymized participant information are provided in Appendix~\ref{app:data_collection}.

The collected and curated \ourdata{} corpus contains approximately \textbf{100 hours} of egocentric video across the seven tool-use domains described above. To balance controlled task coverage with natural tool-use behavior, we adopt two complementary collection settings: \textbf{structured task-guided} and \textbf{open-ended participant-driven}.
In the structured task-guided setting, participants follow predefined task sequences prepared by the data collection team. Each sequence specifies a task goal and key procedural steps, yielding clear task boundaries, observable task-state changes, and controlled coverage of hand--tool--object interaction. In the open-ended participant-driven setting, participants are given only a broad topic or high-level goal and complete the task in their own manner. This setting allows spontaneous tool choices, procedural adaptation, repeated attempts, error recovery, and opportunistic substitutions. Together, the structured setting provides consistent procedural data for reliable annotation and benchmark construction, while the participant-driven setting captures the variability and adaptivity of in-the-wild tool use.

\paragraph{Source-video partition.}
Before constructing the instruction-tuning corpus and benchmark, we partition the recordings at the source-video level into a training pool and a benchmark-reserved pool. The 40.34-hour benchmark pool is held out from all stages of instruction-data construction. Consequently, no clip, caption, narration, synthetic QA, or other annotation derived from a benchmark source video is included in model training.

\subsection{Multimodal Annotation and Curation}
\label{ssec:annotation}

\paragraph{Textual Annotations.}
We provide two complementary textual annotations: \textbf{dense captions} and \textbf{tool-centric narrations}. Dense captions describe visible actions and task progress across temporal scales, while narrations capture participant intent, tool grounding, and tool-use reasoning. For dense captioning, we use Gemini-3-Flash~\cite{gemini3flash2025} in a streaming hierarchical pipeline: each video is split into 5-minute clips, with 32 uniformly sampled frames used to generate a global context caption. Each clip is then captioned sequentially in 5-second chunks, where the first chunk uses the global caption and later chunks use the previous caption as memory for temporal consistency. Chunk captions are further aggregated into 1-minute windows and 5-minute summaries.
In parallel, object recognition on key frames provides visual evidence for post-hoc hallucination checks. Following common practice in egocentric video benchmarks, where narrations serve as weak supervision for first-person activities~\citep{damen2021rescaling,grauman2022ego4d}, we also collect narrations for \ourmethod{}. Unlike prior datasets that encourage broad descriptions of visible actions, our narrations are explicitly tool-centric: collectors who perform the tasks narrate meaningful tool-use events using a reference narration template, emphasizing tool selection, usage intent, and effects on target objects or task states. For selected narrations, collectors annotate 2D grounding points on corresponding keyframes for focal tools or objects chosen for tracking, linking textual mentions to visual instances and providing initialization cues for long-horizon 3D tracking. The dense captions and corrected English narrations are subsequently converted into temporally grounded video instruction examples, providing supervision for action description, procedural summarization, tool selection and substitution, object-state tracking, and next-step prediction. The textual annotation interfaces are shown in Figure~\ref{fig:text-anno}. Additional details on the narration template and examples are provided in Appendix~\ref{app:narration_guidelines}.

\begin{figure*}[t!]
    \centering
    \includegraphics[width=1.\linewidth]{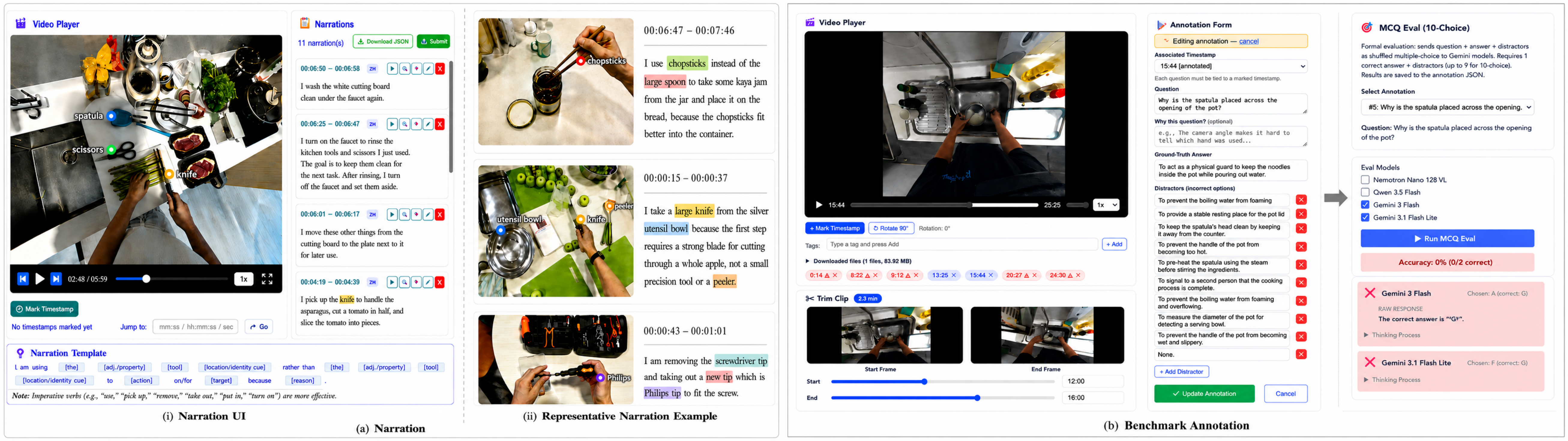}
    \caption{
        \textbf{Textual annotation pipeline.}
        (a) Tool-centric narrations are annotated from grounded egocentric video segments.
        (b) Benchmark QA pairs are constructed and refined through an annotation interface with MLLM-assisted checking.
    }
    \label{fig:text-anno}
    \vspace{-1em}
\end{figure*}

\begin{wrapfigure}{r}{0.6\textwidth}
    \centering
    \vspace{-0.8em}
    \includegraphics[width=0.58\textwidth]{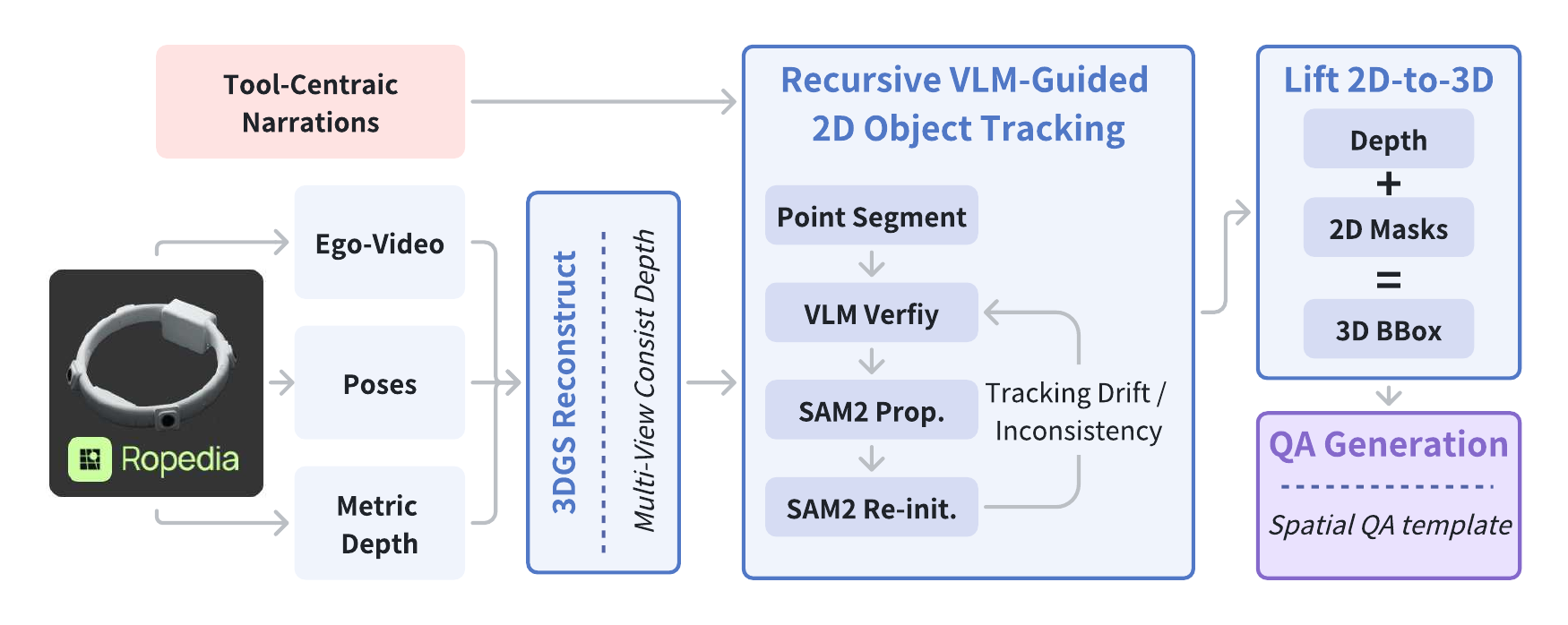}
    \caption{
    \textbf{3D annotation pipeline.}
    Raw geometry from the HOMIE device is reconstructed with 3DGS to obtain consistent depth, then combined with tool-centric narrations for recursive VLM-guided object tracking. The resulting masks are lifted to 3D for object boxes and spatial QA generation.
    }
    \vspace{-1em}
    \label{fig:3d-anno}
\end{wrapfigure}

\paragraph{3D Annotations.}
As shown in Figure~\ref{fig:3d-anno}, we collect raw geometric data with the HOMIE device from Ropedia, Inc., including egocentric videos, camera poses, and depth maps. Following Holi-Spatial~\cite{gao2026holi}, we train a 3D Gaussian Splatting model~\cite{li2024langsurf} to build a multi-view consistent scene geometry, reduce artifacts such as floaters, and render high-fidelity, temporally coherent depth maps for each frame.
\par

\noindent
For robust long-term object tracking, we introduce a recursive VLM-guided segmentation framework. Tool-centric narrations provide initial 2D grounding points and semantic captions, from which SAM2~\cite{ravi2024sam} propagates object masks through the video. When later frames contain additional grounding points, a VLM~\cite{team2024gemini} verifies tracking consistency. If drift or semantic mismatch is detected, the framework re-initializes SAM2 from the failure point and refines the segmentation. This feedback loop yields spatio-temporal masks with both geometric precision and long-range semantic coherence.
Finally, we lift the tracked masks into 3D using the reconstructed depth and camera poses, enabling object boxes and long-horizon 4D object trajectories. Based on these annotations, we design spatial QA templates that query object motion and spatial relations. Together, the original 100-hour egocentric videos, dense textual annotations, and 3D annotations constitute \ourdata{}, from which we derive a video instruction-tuning corpus and a source-video-disjoint \ourbench{} evaluation set.

\subsection{Video Instruction-Tuning Data Construction}
\label{ssec:sft_data}

To evaluate whether \ourdata{} provides actionable training supervision, we construct a video instruction-tuning corpus exclusively from the EgoTools training pool. Every training instance is derived from EgoTools videos and their associated dense captions, corrected English narrations, and temporal metadata.
The instruction-tuning corpus contains \textbf{184,679} examples, and every example is labeled with the capability it supervises. \textbf{Grounding-oriented} supervision asks what is visible and where: which tool is in use, which object it contacts, what attribute or local state has changed, and how hands, tools, and objects are arranged in space. \textbf{Reasoning-oriented} supervision asks why and in what order: why a tool is chosen for a material or goal, what the causal effect of an action is, how a task progresses through its steps, and what happens next. A third group provides \textbf{descriptive} supervision through dense captioning and narration completion, which teaches the model to verbalize first-person visual evidence without a question format.

\begin{figure}[t]
\centering
\begin{minipage}[c]{0.46\linewidth}
  \centering
  \resizebox{\linewidth}{!}{\definecolor{sbToolI}{HTML}{6BAED6}\definecolor{sbToolA}{HTML}{D3E4F3}\definecolor{sbToolB}{HTML}{BDD7EB}\definecolor{sbToolC}{HTML}{A8CCE5}\definecolor{sbToolD}{HTML}{93C0DF}
\definecolor{sbReasonI}{HTML}{E8A33D}\definecolor{sbReasonA}{HTML}{F8D094}\definecolor{sbReasonB}{HTML}{FBE3BC}\definecolor{sbReasonC}{HTML}{FDF1DC}
\definecolor{sbQAI}{HTML}{74C476}    \definecolor{sbQAA}{HTML}{CDE9CD}    \definecolor{sbQAB}{HTML}{E6F4E6}
\definecolor{sbDescI}{HTML}{E78AC3}  \definecolor{sbDescA}{HTML}{F7D3E7}
\definecolor{sbOpenI}{HTML}{9E9AC8}  \definecolor{sbOpenA}{HTML}{C9C6E0}

\newcommand{\sbseg}[5]{\fill[#5,draw=white,line width=1.2pt]
    (#3:#1) arc (#3:#4:#1) -- (#4:#2) arc (#4:#3:#2) -- cycle;}

\newcommand{\sbcurve}[5]{\pgfmathsetmacro{\sbm}{mod((#2+#3)/2,360)}\pgfmathtruncatemacro{\sbup}{ifthenelse(\sbm<180,1,0)}\ifnum\sbup=1\relax
    \path[decorate,decoration={text along path,text align={center},
      raise=-0.55ex,text={|#4|#5}}] (#3:#1) arc (#3:#2:#1);
  \else
    \path[decorate,decoration={text along path,text align={center},
      raise=-0.55ex,text={|#4|#5}}] (#2:#1) arc (#2:#3:#1);
  \fi}

\newcommand{\sblabO}[5]{\pgfmathsetmacro{\sbm}{(#3+#4)/2}\pgfmathsetmacro{\sbe}{ifthenelse(mod(\sbm,360)>90 && mod(\sbm,360)<270, 1, 0)}\draw[gray!65,line width=0.4pt] (\sbm:#1) -- (\sbm:#2);
  \ifnum\sbe=1
    \node[anchor=east,align=right,font=\sffamily\footnotesize,inner sep=1.5pt] at (\sbm:#2) {#5};
  \else
    \node[anchor=west,align=left,font=\sffamily\footnotesize,inner sep=1.5pt] at (\sbm:#2) {#5};
  \fi}

\begin{tikzpicture}[font=\sffamily]
  \def\ra{1.34} \def\rb{2.46} \def\rc{4.06}
  \def\rA{3.62} \def\rB{3.26} \def\rC{2.90}
  \def\ri{2.10}

  \sbseg{\rb}{\rc}{28}{52.76}{sbToolA}          \sbseg{\rb}{\rc}{52.76}{78.53}{sbToolB}       \sbseg{\rb}{\rc}{78.53}{118.61}{sbToolC}      \sbseg{\rb}{\rc}{118.61}{149.86}{sbToolD}     \sbseg{\rb}{\rc}{149.86}{271.36}{sbReasonC}   \sbseg{\rb}{\rc}{271.36}{326}{sbQAA}          \sbseg{\rb}{\rc}{326}{356}{sbDescA}           \sbseg{\rb}{\rc}{356}{388}{sbOpenA}           \par
  \sbseg{\ra}{\rb}{28}{149.86}{sbToolI}
  \sbseg{\ra}{\rb}{149.86}{271.36}{sbReasonI}
  \sbseg{\ra}{\rb}{271.36}{326}{sbQAI}
  \sbseg{\ra}{\rb}{326}{356}{sbDescI}
  \sbseg{\ra}{\rb}{356}{388}{sbOpenI}
  \fill[white] (0,0) circle (\ra);

  \sbcurve{\rA}{28}{52.76}{\sffamily\scriptsize\bfseries}{Affordance}
  \sbcurve{\rB}{28}{52.76}{\sffamily\scriptsize\bfseries}{Causality}
  \sbcurve{\rC}{28}{52.76}{\sffamily\scriptsize}{14.2K}

  \sbcurve{\rA}{52.76}{78.53}{\sffamily\scriptsize\bfseries}{Perception}
  \sbcurve{\rB}{52.76}{78.53}{\sffamily\scriptsize\bfseries}{Grounding}
  \sbcurve{\rC}{52.76}{78.53}{\sffamily\scriptsize}{14.8K}

  \sbcurve{\rA}{78.53}{118.61}{\sffamily\scriptsize\bfseries}{Procedural}
  \sbcurve{\rB}{78.53}{118.61}{\sffamily\scriptsize\bfseries}{Dynamics}
  \sbcurve{\rC}{78.53}{118.61}{\sffamily\scriptsize}{22.9K}

  \sbcurve{\rA}{118.61}{149.86}{\sffamily\scriptsize\bfseries}{Spatial}
  \sbcurve{\rB}{118.61}{149.86}{\sffamily\scriptsize\bfseries}{Reasoning}
  \sbcurve{\rC}{118.61}{149.86}{\sffamily\scriptsize}{17.9K}

  \sbcurve{3.10}{149.86}{271.36}{\sffamily\scriptsize\bfseries}{Episode-Level Multiple Choice}
  \sbcurve{3.44}{149.86}{271.36}{\sffamily\scriptsize}{69.6K}

  \sbcurve{\rC}{271.36}{326}{\sffamily\scriptsize\bfseries}{General}
  \sbcurve{\rB}{271.36}{326}{\sffamily\scriptsize\bfseries}{Video QA}
  \sbcurve{\rA}{271.36}{326}{\sffamily\scriptsize}{31.3K}

  \sbcurve{\rC}{326}{356}{\sffamily\scriptsize\bfseries}{Caption}
  \sbcurve{\rB}{326}{356}{\sffamily\scriptsize\bfseries}{Narration}
  \sbcurve{\rA}{326}{356}{\sffamily\scriptsize}{9.1K}

  \sbcurve{\rA}{356}{388}{\sffamily\scriptsize\bfseries}{Single-Image}
  \sbcurve{\rB}{356}{388}{\sffamily\scriptsize\bfseries}{Open QA}
  \sbcurve{\rC}{356}{388}{\sffamily\scriptsize}{5K}

  \sbcurve{1.90}{28}{149.86}{\sffamily\footnotesize\bfseries\color{white}}{Tool-Centric Reasoning}
  \sbcurve{1.90}{149.86}{271.36}{\sffamily\footnotesize\bfseries\color{white}}{Episode Reasoning}
  \sbcurve{1.90}{271.36}{326}{\sffamily\footnotesize\bfseries\color{white}}{General QA}
  \sbcurve{\ri}{326}{356}{\sffamily\scriptsize\bfseries\color{white}}{Caption}
  \sbcurve{\ri}{356}{388}{\sffamily\scriptsize\bfseries\color{white}}{Open QA}

  \node[align=center,font=\sffamily] at (0,0)
    {\textbf{\normalsize EgoTools}\\[1pt]\textbf{\normalsize Training Data}\\[3pt]\footnotesize\color{black!60} (184,679)};
\end{tikzpicture}}\\[1pt]
  {\footnotesize (a)}
\end{minipage}\hfill
\begin{minipage}[c]{0.51\linewidth}
  \centering
  \resizebox{\linewidth}{!}{\providecolor{sbToolI}{HTML}{6BAED6}\providecolor{sbToolSub}{HTML}{A8CCE5}
\providecolor{sbReasonI}{HTML}{E8A33D}\providecolor{sbQAI}{HTML}{74C476}
\providecolor{sbDescI}{HTML}{E78AC3}\providecolor{sbOpenI}{HTML}{9E9AC8}
\begin{tikzpicture}[font=\sffamily]
  \foreach \p in {10,20,30}{
    \pgfmathsetmacro{\gx}{\p*0.0955}
    \draw[black!12,line width=0.4pt] (\gx,-4.30) -- (\gx,0.30);
    \node[font=\sffamily\tiny,text=black!45,anchor=north] at (\gx,-4.35) {\p\%};}
  \node[font=\sffamily\tiny,text=black!45,anchor=north] at (0,-4.35) {0};
  \foreach \y/\name/\cnt/\nclr/\clr/\len/\bh in {0/{\bfseries Tool-Centric Reasoning}/{\bfseries 69.8K\, (37.8\%)}/black/sbToolI/3.610/0.30,
    -0.46/{Affordance Causality}/{14.2K\, (7.7\%)}/black!65/sbToolSub/0.735/0.22,
    -0.90/{Perception Grounding}/{14.8K\, (8.0\%)}/black!65/sbToolSub/0.764/0.22,
    -1.34/{Procedural Dynamics}/{22.9K\, (12.4\%)}/black!65/sbToolSub/1.184/0.22,
    -1.78/{Spatial Reasoning}/{17.9K\, (9.7\%)}/black!65/sbToolSub/0.926/0.22,
    -2.46/{\bfseries Episode-Level Multiple Choice}/{\bfseries 69.6K\, (37.7\%)}/black/sbReasonI/3.600/0.30,
    -3.08/{\bfseries General Video QA}/{\bfseries 31.3K\, (16.9\%)}/black/sbQAI/1.614/0.30,
    -3.70/{\bfseries Caption \& Narration}/{\bfseries 9.1K\, (4.9\%)}/black/sbDescI/0.468/0.30,
    -4.30/{\bfseries Single-Image Open QA}/{\bfseries 5K\, (2.7\%)}/black/sbOpenI/0.258/0.30}{
    \node[anchor=east,font=\sffamily\scriptsize,text=\nclr,inner sep=1pt] at (-0.15,\y+\bh/2) {\name};
    \fill[\clr,rounded corners=1.2pt] (0,\y) rectangle (\len,\y+\bh);
    \node[anchor=west,font=\sffamily\scriptsize,text=black!60,inner sep=2pt] at (\len+0.05,\y+\bh/2) {\cnt};
  }
  \draw[black!25,line width=0.5pt] (0,-4.30) -- (0,0.30);
\end{tikzpicture}}\\[6pt]
  {\footnotesize (b)}
\end{minipage}
\caption{\textbf{Composition of the EgoTools-Data instruction-tuning corpus.}
(a) The inner ring groups the 184,679 examples by what they supervise, and the outer ring gives the corresponding categories; segment angles are schematic rather than strictly proportional, so that the smallest categories remain readable. (b) Shares drawn to scale; the four tool-centric categories are nested under their group total. Exact numbers are listed in Table~\ref{tab:training_data_composition}.}
\label{fig:train_data_composition}
\end{figure}
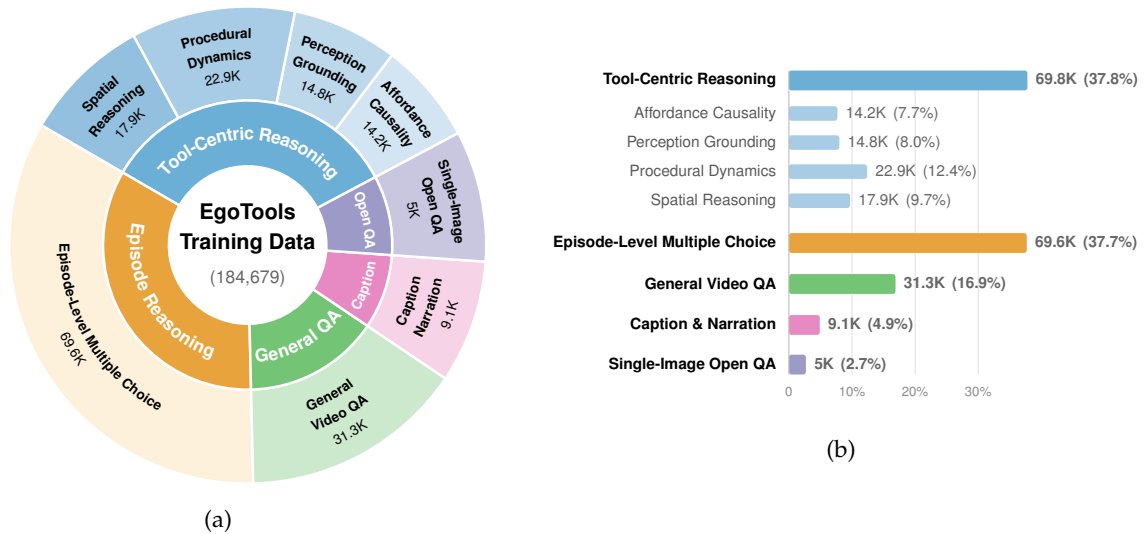

Concretely, 69,760 examples (37.8\%) supervise the four \ourbench{} abilities directly, split into affordance and causality (14,175), perception and grounding (14,755), procedural dynamics (22,943), and spatial reasoning (17,887). A further 69,554 examples (37.7\%) are episode-level multiple-choice questions about action ordering, overall activity, and task goals, including 5,000 next-action questions that pair a video with the current observation frame. General video QA contributes 31,281 examples (16.9\%) of mixed episode-level questions. The remainder is descriptive or open-ended: 9,084 examples (4.9\%) of dense captioning and narration completion, and 5,000 (2.7\%) of single-image open-ended QA. Synthetic QA candidates are filtered for visual answerability, temporal grounding, single-best-answer validity, and resistance to textual shortcuts, and answer letters are balanced within each option-count group to reduce positional bias. Figure~\ref{fig:train_data_composition} summarizes the resulting distribution.
\subsection{Benchmark Construction and Quality Control}
\label{ssec:benchmark_construction}

\paragraph{Benchmark QA Construction.}
Based on \ourdata{}, we select a high-quality \textbf{40.34-hour} benchmark-reserved subset for \ourbench{} construction, prioritizing annotation quality, tool-use density, task diversity, and environmental coverage. The benchmark and instruction-tuning corpus are disjoint at the source-video level. We manually create QA pairs with both original data collectors and external annotators: collectors contribute task-aware procedural and causal questions, while external annotators identify visually inferable but model-challenging reasoning points. Domain-specific videos, such as wet-lab procedures, are assigned to annotators with relevant expertise.
Annotators identify segments requiring physically grounded tool-use reasoning, including tool selection, object-state changes, and temporal ordering of tool-mediated actions, and formulate 8-way multiple-choice questions with one correct answer and seven distractors. Each benchmark clip is extracted as a localized temporal window around an anchor moment, with most clips lasting at least three minutes. We limit temporal overlap and control QA density across source videos to maintain diversity over videos, tools, and task contexts. This yields \textbf{900} human-crafted QA pairs.
We further generate \textbf{100} spatially grounded QA pairs from 3D annotations using the Holi-Spatial~\cite{gao2026holi} pipeline, augmented with focal-tool trajectories and object-state signals. These questions target spatial motion, state changes, and tool--object interaction dynamics, and are included only after human verification. Overall, the average temporal span per QA is \textbf{4.19 minutes}.

\paragraph{Quality Control.}
All benchmark items undergo a fix-first quality-control pipeline that repairs valid annotator intent before removal. We audit structural validity, video grounding, and answer-choice quality to detect malformed questions, duplicates, empty submissions, missing visual evidence, temporal misalignment, inconsistent tool identity, and weak distractors. Empty or verbatim-duplicate items are removed; repairable issues are fixed with deterministic rules or minimal model-assisted edits, including reference normalization, first-person leakage removal, and unambiguous answer completion. Low-confidence cases are sent to human review.
After repair, questions are normalized to an 8-way multiple-choice format and screened for shortcuts such as near-duplicate choices, revealing wording, lexical imbalance, answer-length or position bias, and text-only solvability. Text-only and multi-model sanity checks are used to flag potentially trivial or shortcut-solvable items for review. Flagged items are hardened by replacing distractors with visually plausible, length-balanced alternatives while preserving the original question and ground-truth answer when possible. Final revision and retention decisions are based on visual grounding, unique answerability, and annotation quality rather than the correctness of any particular model. Details are in Appendix~\ref{app:qc}.

\paragraph{Independent Reliability Study.}
To quantitatively assess annotation reliability, we conducted an independent study on a stratified sample of 250 benchmark questions, including 100 Affordance \& Causality questions. Five annotators who had not written the original questions each evaluated 100 assigned items, such that every question received two independent judgments. Annotators viewed only the video, question, and answer choices without access to the original answer key. Across 500 judgments, agreement with the original key was \textbf{89.8\%}, including 86.5\% on Affordance \& Causality questions. Exact agreement between the two assigned annotators was \textbf{88.4\%} at the item level, and \textbf{94.4\%} of items were judged to have a uniquely best answer supported by visible evidence. Fourteen items (\textbf{5.6\%}) were flagged as ambiguous or having multiple plausible answers; all were manually reviewed, and items failing the unique-answer criterion were revised or replaced. All results reported in this paper were recomputed on the resulting quality-controlled benchmark.

\subsection{Distribution}
\label{ssec:distribution}

Figure~\ref{fig:data_stat} summarizes the composition of \ourbench{}, which contains \textbf{1,000 QA pairs} across four tracks: \textbf{1) Affordance \& Causality} (\textbf{363}), \textbf{2) Perception \& Grounding} (\textbf{236}), \textbf{3) Procedural Dynamics} (\textbf{222}), and \textbf{4) Spatial Reasoning} (\textbf{179}). These tracks cover core aspects of egocentric tool-use understanding, including tool recognition, object-state grounding, hand--tool--object relations, task progress, and the causal effects of tool use. This distribution reflects the goal of \ourbench{}: evaluating tool use as a reasoning problem involving tool selection, application, and effects, while retaining coverage of perception, procedure, and spatial reasoning. Each track is further divided into fine-grained subtracks for detailed analysis of model strengths and failure modes.
The seven collection domains characterize the coverage of the full EgoTools-Data corpus rather than a domain-balanced benchmark. EgoTools-Bench exhibits a strongly long-tailed distribution across domains, with the majority of questions drawn from Kitchen videos; complete domain-level statistics are provided in Appendix~E.1. We therefore focus our primary analysis on reasoning capabilities rather than domain-level performance.
To avoid over-representing individual recordings, we limit temporal overlap among benchmark clips and cap QA density per source video. Figure~\ref{fig:data_stat} also compares the vocabulary distribution of \ourbench{} with EPIC-KITCHENS-100 and Ego4D. The word clouds qualitatively illustrate that \ourbench{} places greater lexical emphasis on tools, manipulated objects, actions, and state changes than on general egocentric activity recognition.

\section{Experiments}
\label{sec:exp}

\subsection{Experimental Setup}
\label{ssec:exp_setup}

We evaluate \ourbench{} with human experts, proprietary models, and open-source models, and compare model performance with existing egocentric video benchmarks including EgoSchema~\cite{mangalam2023egoschema}, EgoPlan~\cite{chen2024egoplanbench}, and EgoThink~\cite{cheng2024egothink}. The model set covers standard video-language models, audio-capable omni models, and thinking-style variants from Gemini~\cite{team2024gemini}, Qwen~\cite{bai2025qwen3vltechnicalreport,bai2025qwen25vltechnicalreport,xu2025qwen25omnitechnicalreport,wang2024qwen2vlenhancingvisionlanguagemodels}, InternVL~\cite{wang2025internvl35advancingopensourcemultimodal,zhu2025internvl3exploringadvancedtraining}, LLaVA~\cite{li2024llavaonevisioneasyvisualtask,zhang2025llavavideovideoinstructiontuning}, MiMo-VL~\cite{coreteam2025mimovltechnicalreport}, and GLM~\cite{vteam2026glm45vglm41vthinkingversatilemultimodal} families. For \ourbench{}, we report overall accuracy and track-wise accuracy following the four tracks introduced in Section~\ref{ssec:distribution}; existing benchmark results are reported when available under comparable settings. For each model, we report the number of input frames and whether audio is used. Most open-source instruct models use 64 uniformly sampled frames; thinking models use 64 or 512 frames depending on their supported setting; Gemini~\cite{team2024gemini} models use 1 FPS video input. Audio-enabled models receive only synchronized non-narration audio. Corrected narrations, narration audio, dense captions, and annotation metadata are never provided as input.

\paragraph{Human evaluation.}
Human performance was measured using five evaluators who had not authored the questions they evaluated. Each evaluator answered all 1,000 benchmark questions over multiple sessions and could freely seek and replay the videos. Evaluators did not have access to the answer key, and there were no skipped or invalid responses. Overall and track-wise accuracies are micro-averaged over 5,000 individual judgments. The human results therefore reflect a free-viewing condition rather than the fixed-frame input budget used for model evaluation.

\paragraph{EgoTools training.}
To evaluate the training utility of \ourdata{}, we adapt Qwen3-VL-8B-Instruct with a single stage of full-parameter supervised fine-tuning of its language-model component, while keeping the visual encoder and multimodal aligner frozen. We train on the 184,679-example EgoTools instruction corpus described in Section~\ref{ssec:sft_data}. No videos, images, or QA annotations from EgoSchema, EgoPlan-Bench, or EgoThink are used, and all training videos are disjoint from \ourbench{} at the source-video level. Optimization and input settings are provided in Appendix~\ref{app:training_details}.

\subsection{Main Results}
\label{ssec:main_results}

\begin{table*}[t]
\centering
\caption{\textbf{Performance comparison across egocentric video understanding benchmarks.}
We compare human performance, proprietary models, and open-source video-language models on three existing egocentric benchmarks and our \ourbench{} benchmark. The last five columns report results on \ourbench{} across four research-facing tracks and the overall average. Track abbreviations are: \textbf{AC} = Affordance \& Causality, \textbf{PG} = Perception \& Grounding, \textbf{PD} = Procedural Dynamics, and \textbf{SR} = Spatial Reasoning.}
\label{tab:main_results}

\resizebox{\textwidth}{!}{\begin{tabular}{lcccccccccc}
\toprule
\multirow{2}{*}{\textbf{Model}}
& \multirow{2}{*}{\textbf{Frames}}
& \multirow{2}{*}{\textbf{Audio}}
& \multirow{2}{*}{\textbf{EgoSchema~\cite{mangalam2023egoschema}}}
& \multirow{2}{*}{\textbf{EgoPlan~\cite{chen2024egoplanbench}}}
& \multirow{2}{*}{\textbf{EgoThink~\cite{cheng2024egothink}}}
& \multicolumn{5}{c}{\textbf{EgoTools}} \\
\cmidrule(lr){7-11}
& & & & &
& \textbf{AC} & \textbf{PG} & \textbf{PD} & \textbf{SR} & \textbf{Overall} \\
\midrule

\modelgroup{Human Baseline} \\
\midrule
Human Expert
& -- & \cmark
& $\sim$76 & -- & --
& 82.1 & 85.2 & 83.3 & 82.7 & 83.2 \\

\midrule
\modelgroup{Proprietary Models} \\
\midrule
Gemini-3.1-Pro~\cite{team2024gemini} & 1fps & \cmark
& -- & -- & --
& \best{69.7} & \best{51.7} & 72.1 & \best{74.9} & \best{66.9} \\

Gemini-3-Flash~\cite{gemini3flash2025}
& 1fps & \cmark
& -- & -- & --
& 64.5 & 50.0 & \best{73.0} & 72.6 & 64.4 \\

Gemini-3.1-Flash-Lite~\cite{gemini31flashlite2026}
& 1fps & \cmark
& -- & -- & --
& 58.1 & 44.5 & 59.9 & 58.7 & 55.4 \\

\midrule
\modelgroup{Open-Source Models (Instruct)} \\
\midrule
Qwen3-VL-8B-Instruct~\cite{bai2025qwen3vltechnicalreport}
& 64 & \xmark
& \second{69.0} & \best{42.3} & \second{61.0}
& \second{46.1} & \best{45.9} & \best{57.1} & \best{55.4} & \best{50.0} \\

Qwen3-VL-4B-Instruct~\cite{bai2025qwen3vltechnicalreport}
& 64 & \xmark
& \best{69.6} & \second{42.0} & \best{61.9}
& 41.7 & \second{45.4} & 50.2 & \third{50.0} & 45.7 \\

Qwen2.5-VL-7B-Instruct~\cite{bai2025qwen25vltechnicalreport}
& 64 & \xmark
& 61.0 & 32.0 & 57.7
& 40.1 & 43.3 & 50.2 & 47.8 & 44.3 \\

Qwen2.5-Omni-7B-Instruct~\cite{xu2025qwen25omnitechnicalreport}
& 64 & \cmark
& \third{65.2} & -- & 58.9
& 44.7 & \best{45.9} & \third{51.8} & 46.7 & \third{47.1} \\

Qwen2-VL-7B-Instruct~\cite{wang2024qwen2vlenhancingvisionlanguagemodels}
& 64 & \xmark
& 63.6 & \third{39.9} & 60.1
& \best{46.6} & 35.6 & 46.2 & 47.8 & 44.2 \\

InternVL3.5-8B-Instruct~\cite{wang2025internvl35advancingopensourcemultimodal}
& 64 & \xmark
& 63.8 & -- & 56.0
& \third{45.8} & \best{45.9} & \second{53.4} & \second{53.3} & \second{48.7} \\

InternVL3-8B-Instruct~\cite{zhu2025internvl3exploringadvancedtraining}
& 64 & \xmark
& \best{69.6} & -- & 59.4
& 44.7 & \third{44.9} & \third{51.8} & 48.9 & \third{47.1} \\

LLaVA-OneVision-7B~\cite{li2024llavaonevisioneasyvisualtask}
& 64 & \xmark
& 63.4 & -- & 54.9
& 37.6 & 37.1 & 42.9 & 37.0 & 38.9 \\

LLaVA-Video-7B-Qwen2~\cite{zhang2025llavavideovideoinstructiontuning}
& 64 & \xmark
& 52.4 & -- & 58.0
& 38.2 & 39.7 & 43.3 & 37.0 & 39.8 \\

MiMo-VL-7B-SFT~\cite{coreteam2025mimovltechnicalreport}
& 64 & \xmark
& 54.4 & 35.9 & \third{60.7}
& 41.4 & \third{44.9} & 49.4 & 40.2 & 44.2 \\

\midrule
\modelgroup{Open-Source Models (Thinking)} \\
\midrule
Qwen3-VL-8B-Thinking~\cite{bai2025qwen3vltechnicalreport}
& 512 & \xmark
& \best{70.3} & \best{43.6} & \third{62.0}
& \second{43.6} & 41.8 & 48.2 & \best{55.4} & \second{45.7} \\

Qwen3-VL-4B-Thinking~\cite{bai2025qwen3vltechnicalreport}
& 512 & \xmark
& \third{59.8} & \second{43.1} & \second{62.3}
& 37.6 & 36.6 & 36.4 & 41.3 & 37.4 \\

MiMo-VL-7B-RL~\cite{coreteam2025mimovltechnicalreport}
& 64 & \xmark
& 57.0 & 36.6 & 59.4
& 43.1 & \second{45.9} & \second{50.6} & 39.1 & 45.3 \\

GLM4.1V-Thinking~\cite{vteam2026glm45vglm41vthinkingversatilemultimodal}
& 64 & \xmark
& \second{66.1} & 25.3 & \best{62.7}
& \best{46.1} & \best{50.0} & \best{57.1} & \second{53.3} & \best{50.7} \\

\midrule
\modelgroup{Ours} \\
\midrule
\textbf{EgoTools-8B (Ours)}
& 64 & \xmark
& 67.6 & 35.2 & 64.0
& 55.4 & 59.7 & 68.1 & 44.4 & 60.9 \\

\bottomrule
\end{tabular}}
\vspace{-0.5em}
\end{table*}

\paragraph{EgoTools-Bench is challenging for current video-language models.}
Table~\ref{tab:main_results} summarizes performance across human experts, proprietary models, open-source instruct models, and open-source thinking models. EgoTools-Bench uses an 8-way multiple-choice format, so chance performance is 12.5\%. Across open-source instruct models, performance remains low, with overall accuracy ranging from 38.9\% to 50.0\%, far below the human expert score of 83.2\%. The strongest open-source instruct result is Qwen3-VL-8B-Instruct~\cite{bai2025qwen3vltechnicalreport} at 50.0\%, followed by InternVL3.5-8B-Instruct~\cite{wang2025internvl35advancingopensourcemultimodal} at 48.7\% and Qwen2.5-Omni-7B-Instruct~\cite{xu2025qwen25omnitechnicalreport} at 47.1\%. Among open-source thinking models, GLM4.1V-Thinking~\cite{vteam2026glm45vglm41vthinkingversatilemultimodal} achieves the strongest overall result at 50.7\%. These results indicate that EgoTools-Bench exposes a substantial gap in current models' ability to reason about tool-mediated actions, rather than an isolated weakness of one architecture. The final row reports \ourmodel{}-8B, our model fine-tuned on \ourdata{}; it is listed separately from the zero-shot model baselines and analyzed in \S\ref{ssec:sft_results}.

\paragraph{Performance across egocentric benchmarks.}
Table~\ref{tab:main_results} also reports results on existing egocentric benchmarks to contextualize the evaluated models. For example, Qwen3-VL-8B-Instruct reaches 69.0\% on EgoSchema~\cite{mangalam2023egoschema} and 61.0\% on EgoThink~\cite{cheng2024egothink}, compared with 50.0\% on EgoTools-Bench. Qwen3-VL-4B-Instruct shows a similar pattern, with 69.6\% on EgoSchema, 61.9\% on EgoThink, and 45.7\% on EgoTools-Bench. Thus, broad first-person activity understanding does not necessarily imply robust tool-use reasoning about affordances, grounding, procedures, spatial relations, and state changes.

\paragraph{Additional model and input capacity do not close the gap.}
Overall accuracy varies across models with different parameter scales, audio support, and thinking-style inference. We examine these differences by track and model class in §\ref{ssec:analysis}, then by tool-use cues and error patterns in §\ref{ssec:ablation_error}. These results motivate a complementary question: whether the missing tool-centric capabilities can be learned from the supervision provided by \ourdata{}.

\subsection{Training Utility of EgoTools-Data}
\label{ssec:sft_results}

\paragraph{EgoTools supervision improves tool-centric reasoning.}
Under the standard 64-frame MP4-based evaluation protocol, single-stage full supervised fine-tuning on 184,679 EgoTools-derived instruction examples improves Qwen3-VL-8B-Instruct from \textbf{50.0\% to 60.9\%} on \ourbench{}, a gain of 10.9 percentage points. The improvement covers three of the four tracks: +9.3 points on Affordance \& Causality, +13.8 on Perception \& Grounding, and +11.0 on Procedural Dynamics, while Spatial Reasoning decreases by 11.0 points. Because the training and evaluation resources are separated at the source-video level, these gains reflect generalization to held-out tool-use episodes rather than memorization of benchmark clips. The result demonstrates that \ourdata{} is not merely an intermediate source for benchmark construction, but provides actionable supervision for learning egocentric tool-use reasoning. Despite explicit spatial supervision, SR performance decreases after fine-tuning, indicating that spatial reasoning remains a limitation of the current training recipe.

\paragraph{Performance on EgoSchema.}
The single-stage model reaches \textbf{67.6\%} on EgoSchema, 1.4 points below the 69.0\% backbone.

\paragraph{The EgoPlan and EgoThink results are consistent with answer-format coverage.}
Beyond multiple-choice video QA, the training mixture covers two further answer formats built from EgoTools training videos: 5,000 next-action multiple-choice examples that pair a video with the current observation frame, and 5,000 single-image open-ended examples. These formats match how EgoPlan-Bench and EgoThink pose their questions. On EgoThink, the model reaches \textbf{64.0\%} macro accuracy, exceeding the 61.0\% backbone by 3.0 points, whereas on EgoPlan-Bench it reaches 35.2\%, still 7.1 points below the 42.3\% backbone. Because these training subsets mirror the answer formats used by EgoPlan-Bench and EgoThink, these results do not isolate the effects of format coverage from other effects of fine-tuning. We therefore limit our primary claim to the training utility of \ourdata{} for tool-centric reasoning rather than claiming uniform improvement across all egocentric benchmarks.

\begin{table*}[t]
\centering
\caption{
\textbf{Training utility of EgoTools-Data.}
\ourmodel{}-8B is Qwen3-VL-8B-Instruct after a single stage of full supervised fine-tuning on 184,679 EgoTools-derived instruction examples. On \ourbench{}, both models use the same standard 64-frame MP4-based evaluation protocol. The instruction-tuning corpus and \ourbench{} are disjoint at the source-video level. \ourbench{} results are reported on the full quality-controlled 1,000-question benchmark; EgoSchema is reported on the full 5,031-question set, EgoPlan-Bench on its 3,343-question validation set, and EgoThink as the macro-averaged score over its twelve single-image subtasks.
}
\label{tab:sft_results}
\resizebox{\textwidth}{!}{\begin{tabular}{lccccc|ccc}
\toprule
\multirow{2}{*}{\textbf{Model}}
& \multicolumn{5}{c|}{\textbf{EgoTools-Bench}}
& \multicolumn{3}{c}{\textbf{Existing Egocentric Benchmarks}} \\
\cmidrule(lr){2-6}
\cmidrule(lr){7-9}
& \textbf{AC} & \textbf{PG} & \textbf{PD} & \textbf{SR}
& \textbf{Overall}
& \textbf{EgoSchema}
& \textbf{EgoPlan}
& \textbf{EgoThink} \\
\midrule
Qwen3-VL-8B-Instruct
& 46.1 & 45.9 & 57.1 & \textbf{55.4} & 50.0
& \textbf{69.0} & \textbf{42.3} & 61.0 \\

\textbf{EgoTools-8B}
& \textbf{55.4} & \textbf{59.7} & \textbf{68.1} & 44.4 & \textbf{60.9}
& 67.6 & 35.2 & \textbf{64.0} \\
\bottomrule
\end{tabular}}
\end{table*}

\subsection{Analysis}
\label{ssec:analysis}

\paragraph{Perception \& Grounding is the diagnostic anchor.}
Perception \& Grounding (PG) is particularly dependent on concrete visual evidence because it asks about the focal tool, target object, contact point, object attribute, quantity, or local state change. Activity context alone rarely determines which visually similar tool is being used, which object it contacts, or what has changed locally. Open-source instruct models score between 35.6\% and 45.9\% on PG, and even Gemini-3.1-Pro reaches 51.7\%, still 18--23 points below its AC, PD, and SR scores. Together with the input ablation in Table~\ref{tab:input_ablation}, where PG shows the largest improvement from visual evidence over text-only input, these results identify precise visual grounding as a persistent bottleneck.

\paragraph{Model variants exhibit different track-wise strengths.}
The higher-level tracks admit different sources of contextual support: tool-function priors for AC, task-order context for PD, and workspace geometry and physical commonsense for SR. Qwen3-VL-8B-Instruct improves over its 4B counterpart most strongly on PD and SR, by 6.9 and 5.4 points, respectively, while improving PG by only 0.5 points. For this model pair, the performance difference is larger on procedural and spatial reasoning than on precise grounding. The omni and thinking variants also show mixed patterns: Qwen2.5-Omni-7B-Instruct improves over Qwen2.5-VL-7B-Instruct on AC, PG, and PD but scores slightly lower on SR, while Qwen3-VL-8B-Thinking underperforms its instruct counterpart on AC, PG, and PD and only matches it on SR. In contrast, GLM4.1V-Thinking achieves the strongest zero-shot open-source result in Table~\ref{tab:main_results}, at 50.7\%. These heterogeneous patterns characterize the evaluated configurations; because the variants differ in training and sometimes input budgets, they do not isolate the effects of scale, audio, or deliberation.

\subsection{Ablations and Error Analysis}
\label{ssec:ablation_error}

\begin{table}[t]
\centering
\small
\setlength{\tabcolsep}{4.5pt}
\renewcommand{\arraystretch}{1.05}
\caption{
\textbf{Input-evidence ablation on the full 1,000-question EgoTools-Bench.}
All conditions use the same Qwen3-VL-8B-Instruct backbone, deterministic decoding, and scoring configuration. The visual conditions use the same pre-extracted frame-list pipeline, with the ordered condition serving as the matched within-protocol control.
}
\label{tab:input_ablation}

\begin{tabular}{lccccc}
\toprule
\textbf{Input condition}
& \textbf{Overall}
& \textbf{AC}
& \textbf{PG}
& \textbf{PD}
& \textbf{SR} \\
\midrule
Ordered 64 frames
& \textbf{51.7}
& \textbf{47.4}
& \textbf{50.0}
& \textbf{57.1}
& \textbf{57.6} \\
Shuffled 64 frames
& 49.3
& 45.5
& 47.4
& 53.9
& 56.5 \\
Middle frame
& 46.9
& 44.7
& 45.9
& 51.0
& 45.7 \\
Text-only
& 42.7
& 39.0
& 36.1
& 51.4
& 47.8 \\
\bottomrule
\end{tabular}
\end{table}

\paragraph{EgoTools-Bench requires visual evidence beyond language priors.}
Table~\ref{tab:input_ablation} compares temporally ordered 64-frame input with the same frames in shuffled order, a single middle frame, and text-only input on the full 1,000-question benchmark. All conditions yield zero parsing or inference failures. Ordered video achieves 51.7\% overall accuracy, compared with 42.7\% for text-only input, corresponding to a 9.0-point improvement. The substantial gain from visual input confirms that the benchmark cannot be solved solely through answer-choice patterns, tool-function commonsense, or language priors.
Text-only performance nevertheless remains above the 12.5\% chance level. This indicates that some questions retain exploitable commonsense or answer-option cues, despite the benchmark's anti-shortcut filtering. We therefore interpret text-only performance as evidence of residual language priors rather than as evidence that video input is unnecessary.

\paragraph{Multi-frame coverage is important.}
Using only the middle frame reduces overall accuracy from 51.7\% to 46.9\%, showing that a single static observation is insufficient for many questions. The 4.8-point gap indicates that models benefit from observing multiple moments of the tool-use process, including the selection of a tool, its interaction with a target object, intermediate state changes, and the resulting task outcome.
The effect is particularly strong for Spatial Reasoning, where ordered multi-frame input outperforms the middle-frame condition by 11.9 points. This result suggests that spatial questions frequently require integrating evidence distributed across the clip rather than reading a single hand--tool--object configuration.

\paragraph{Temporal order provides additional information.}
Shuffling the same 64 frames reduces overall accuracy from 51.7\% to 49.3\%. Because the visual content and frame count remain unchanged, the 2.4-point difference isolates the contribution of temporal ordering. Procedural Dynamics is the most order-sensitive track, with ordered input outperforming shuffled input by 3.2 points. This is consistent with the track's emphasis on step transitions, action ordering, preparation, and changes in tool use over time.
By contrast, Spatial Reasoning shows only a 1.1-point difference between ordered and shuffled frames. Together with its large ordered-versus-middle-frame gap, this pattern indicates that SR benefits primarily from multi-frame coverage, while depending less strongly on the precise temporal sequence of those frames.

\paragraph{The benchmark tracks require complementary evidence.}
The ablation effects differ systematically across the four tracks. Perception \& Grounding shows the largest dependence on visual evidence, improving by 13.9 points over text-only input. Procedural Dynamics is the most sensitive to temporal order, while Spatial Reasoning benefits most from multi-frame coverage. These complementary patterns support the intended diagnostic roles of the benchmark tracks: PG tests grounding in concrete visual evidence, PD tests temporally ordered process understanding, and SR tests the integration of spatial evidence across multiple observations.
All ablation conditions use the same deterministic frame-list pipeline. The ordered condition therefore serves only as the matched control for comparisons within this ablation. Its absolute score is not directly compared with the separate main benchmark result obtained using the standard MP4-based evaluation protocol.

\Needspace{14\baselineskip}
\begin{wraptable}[11]{r}{0.58\linewidth}
\vspace{-4pt}
\centering
\scriptsize
\setlength{\tabcolsep}{4.5pt}
\renewcommand{\arraystretch}{1.08}
\caption{\textbf{Performance on implicit and explicit tool-use questions.}
$\Delta$ denotes Explicit$-$Implicit.}
\label{tab:explicit_implicit_ablation}
\resizebox{\linewidth}{!}{\begin{tabular}{@{}lc|cc|c@{}}
\toprule
\textbf{Model}
& \textbf{Full Bench.}
& \textbf{Implicit}
& \textbf{Explicit}
& \textbf{$\Delta$} \\
\midrule
Gemini-3.1-Pro
& \textbf{66.9}
& \textbf{65.2}
& \textbf{72.9}
& \gain{+7.7} \\

Gemini-3-Flash
& 64.4
& 63.1
& 68.0
& \gain{+4.9} \\

Gemini-3.1-Flash-Lite
& 55.4
& 54.3
& 59.5
& \gain{+5.2} \\

\midrule
Qwen3-VL-8B-Instruct
& 50.0
& 49.1
& 53.1
& \gain{+4.0} \\

Qwen3-VL-8B-Thinking
& 45.7
& 45.8
& 45.2
& \drop{-0.6} \\

\bottomrule
\end{tabular}}
\vspace{-2pt}
\end{wraptable}

\paragraph{Implicit tool-use questions are generally harder.}
Table~\ref{tab:explicit_implicit_ablation} separates questions that explicitly mention the relevant tool-use cue from questions where the model must infer it from the video context. Proprietary Gemini models show a consistent advantage on explicit questions, with explicit-minus-implicit gaps of 4.9--7.7 points. Qwen3-VL-8B-Instruct shows the same direction with a smaller 4.0-point gap. Because the explicit and implicit subsets contain different questions, these gaps may also reflect differences in their content and difficulty. Qwen3-VL-8B-Thinking is the exception, with a small negative gap of 0.6 points. Implicit questions provide a useful subset for evaluating tool-use understanding when the relevant cue is not explicitly stated in the question.

\begin{wrapfigure}{r}{0.45\textwidth}
    \centering
    \vspace{-0.8em}
    \includegraphics[width=0.43\textwidth]{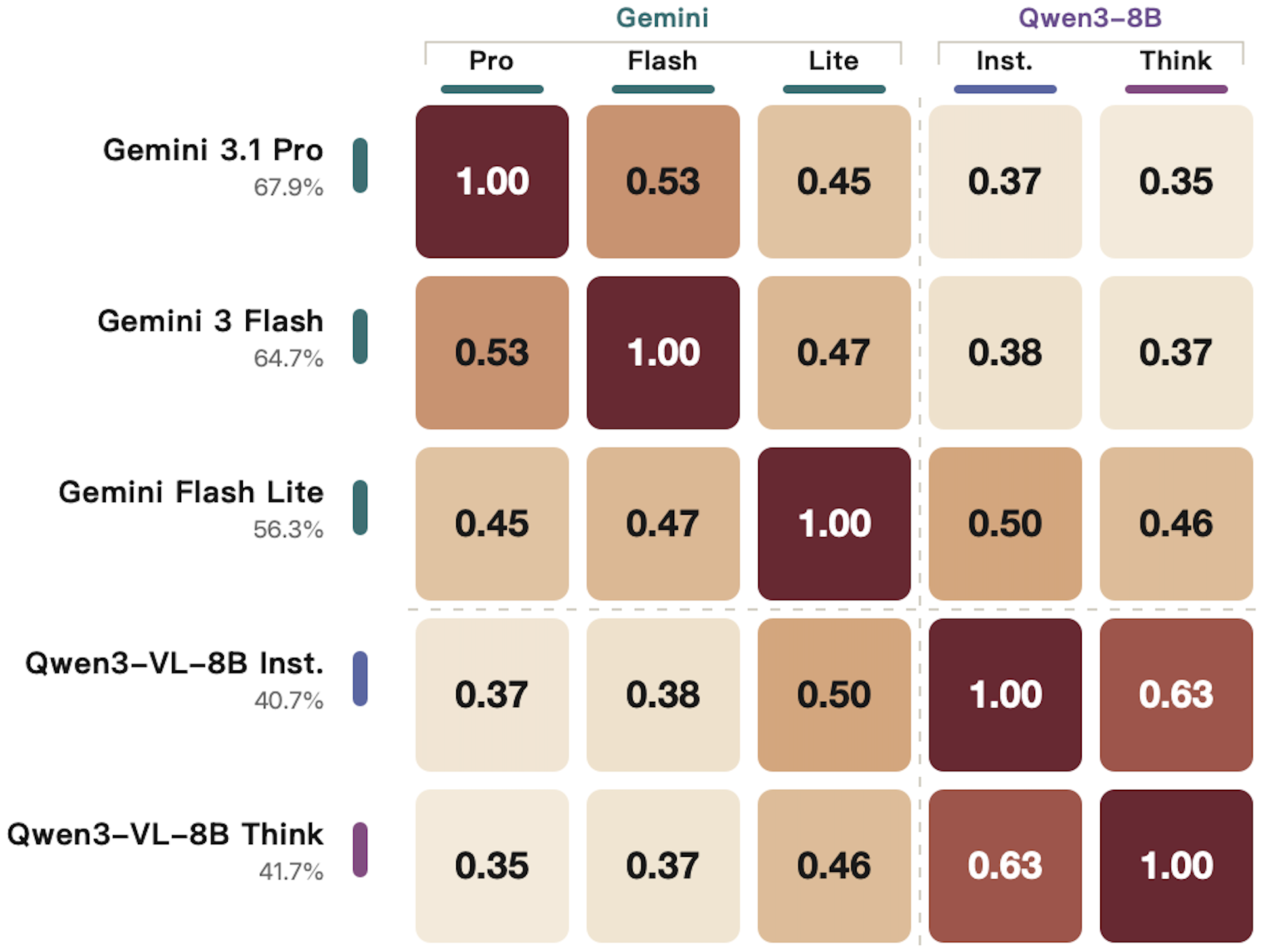}
    \caption{
    \textbf{Jaccard overlap of model errors on EgoTools-Bench.} This highlights shared failure patterns within and across model families.
    }
    \vspace{-1em}
    \label{fig:error-map}
\end{wrapfigure}

\paragraph{Error overlap reveals shared failures.}
The error map in Figure~\ref{fig:error-map} compares the Jaccard overlap between models' incorrect predictions. The strongest off-diagonal overlap is between Qwen3-VL-8B-Instruct and Qwen3-VL-8B-Thinking (0.63), showing substantial shared errors between these two variants. Gemini models also share a moderate error profile with one another, especially Pro and Flash (0.53), but their overlap with Qwen3-VL-8B is lower (0.35--0.38 for Pro/Flash). Gemini-3.1-Flash-Lite has higher overlap with Qwen3-VL-8B-Instruct (0.50) and Qwen3-VL-8B-Thinking (0.46). Overall, the overlap pattern supports that failures are systematic and partially model-family-specific: thinking-style variants retain many failures of their instruct counterparts, while proprietary and open-source models exhibit more complementary error sets.
Taken together, the track-wise results, input ablations, implicit-explicit comparison, and error-overlap analysis provide complementary views of model performance on EgoTools-Bench. These analyses help diagnose weaknesses in tool-use understanding across reasoning tracks, input conditions, and question subsets.

\section{Conclusion}
\label{sec:conclusion}
We introduce EgoTools, a tool-centric egocentric video suite that reframes embodied video understanding around the physical instruments through which humans act, adapt, and transform the world. By combining real-world tool-use recordings, dense hierarchical annotations, long-horizon spatial grounding, and a diagnostic QA benchmark, EgoTools exposes a key limitation of current multimodal video models: recognizing activities is not sufficient to understand how tools mediate action, causality, procedure, and state change. At the same time, supervised fine-tuning on EgoTools-derived instruction data improves Qwen3-VL-8B-Instruct from 50.0\% to 60.9\% on EgoTools-Bench, demonstrating that the proposed data provides actionable supervision for tool-centric reasoning. The gains span three of the four tracks, while Spatial Reasoning remains a limitation of the current training recipe. Nevertheless, a substantial gap to human performance remains, particularly in grounding tool-mediated interactions in precise visual evidence. Ultimately, EgoTools pushes egocentric video understanding beyond recognizing what people do toward understanding how intelligence operates through tools, opening a path toward embodied AI systems that can observe, reason, and assist in the physical world.

{
\small
\bibliography{references}
}

\newpage
\appendix
  \startcontents[appendix]
  \printcontents[appendix]{l}{1}{\section*{Appendix Contents}\vspace{-0.5em}}
  \newpage

\section{Contributors and Acknowledgments}
\label{app:contributors}

\paragraph{Author contributions.}
\textbf{Shulin Tian, Junsu Kim, Shuai Liu,} and \textbf{Hao Li} contributed equally, mainly to data collection, benchmark construction, training, evaluation, and writing. \textbf{Yujiao Shen, Sihan Li, Zhe Yang, Yeongon Kim, Feiyu Li, Jialin Wu, Yichi Zhang, Wenhui Wang, Runmao Yao,} and \textbf{Yuhao Dong} contributed to data annotation and logistics. \textbf{Zhaoxi Chen, Fangzhou Hong, Ziwei Liu, Antonino Furnari, Jingkang Yang,} and \textbf{Hongyuan Zhu} supervised the project and guided its overall research direction. \textbf{Ziwei Liu} served as the corresponding author.

\paragraph{Acknowledgments.}
This work is supported by the Ministry of Education, Singapore, under its MOE AcRF Tier 2 grant (MOE-T2EP20223-0002), and by cash and in-kind funding from NTU S-Lab and its industry partners. We thank Ropedia for providing hardware, API access, and annotation support. We are grateful to \textbf{Jewoo Park, Jeongwon Yoo, Yulgyeom Kim, Yuan Fei, Haiheng Liu, Gordon Chen, Jiarui Zheng, Junhan Zhu,} and \textbf{Hongmei Xu} for their assistance with egocentric video data collection and annotation. We also thank all participants, annotators, reviewers, and infrastructure contributors for their efforts in data collection, annotation, quality control, and release preparation for EgoTools.
\section{Ethics}
\label{app:ethics}

EgoTools was constructed with attention to informed participation, permitted data use, privacy protection, and responsible public release. We provide an anonymized summary of the consent, data-use, and release safeguards applied during data collection, annotation, benchmark construction, model training, and dataset distribution.

\paragraph{Voluntary participation and informed consent.}
Participation in egocentric video recording, narration, and annotation was voluntary. Participants received no monetary compensation for contributing recordings, narrations, or annotations. Before contributing data, participants were informed about the purpose of the project, the types of data to be collected, the intended research uses, and the possibility that cleared materials would be publicly released for research purposes. Participants provided consent covering the applicable data collection, annotation, research-use, and release activities.

\paragraph{Institutional ethics review.}
No formal IRB or institutional ethics approval was sought for the EgoTools data collection. We explicitly report the absence of formal ethics review and describe below the informed-consent, privacy, anonymization, and release safeguards applied throughout the project.

\paragraph{Consent and permitted use.}
Participants contributed egocentric videos, narrations, and associated annotations for research on physical tool-use reasoning, egocentric video understanding, benchmark construction, grounded video-language understanding, multimodal model training, and dataset release. Within the applicable consent and data-use scope, derived materials may include corrected English narrations, dense hierarchical captions, benchmark question--answer pairs, timestamps, grounding annotations, object trajectories, and other research annotations generated from the contributed recordings.

\paragraph{Collected data.}
Depending on the participant role and recording setup, internally collected materials may include raw fisheye videos, rectified egocentric videos, synchronized audio, spoken narrations, corrected English text narrations, dense hierarchical captions, benchmark QA annotations, clip metadata, timestamp annotations, grounding-oriented focal-tool annotations, depth or pose information, and selected trajectory or object-state annotations.

\paragraph{Public release scope.}
The public release is restricted to materials covered by the applicable participant consent, release authorization, anonymization procedures, and privacy review. The release focuses on rectified egocentric videos, dense hierarchical captions, corrected English narrations, benchmark QA pairs, clip metadata, timestamps, split information, and selected finalized grounding annotations. Original narration audio and selected 3D trajectory assets are released only when covered by the corresponding release authorization and privacy review. Raw fisheye videos are not publicly released.

\paragraph{Privacy and release safeguards.}
Released metadata excludes names, contact information, institution-specific identifiers, public profiles, and exact per-video contribution counts. Sensitive, private, or accidentally captured content is reviewed before release, and materials deemed unsuitable for public distribution are excluded. Additional grounding-oriented annotations and 3D trajectory assets are released incrementally only after annotation finalization and the corresponding authorization and privacy checks.

\paragraph{Participant rights and data removal.}
Before public release, participants may request the exclusion of data associated with their participation in accordance with the project release policy. Requests received after release are handled according to the applicable consent terms, dataset license, and release procedures.

\paragraph{Research-use restrictions.}
EgoTools is distributed under a research-use license specifying permitted uses, redistribution conditions, and usage restrictions. The dataset is not intended for participant identification, biometric profiling, surveillance, or other uses that conflict with the original consent and release scope.

\section{Dataset Collection and Contributor Information}
\label{app:data_collection}

\subsection{Data Collection Settings}
\label{app:collection_examples}

EgoTools uses two complementary collection settings to balance procedural control with natural variation in real-world tool use. We provide representative anonymized examples below.

\paragraph{Structured task-guided recording.}
In a structured task-guided session, the data collection team prepares a task goal and a sequence of key procedural steps, while allowing the participant to perform the manipulation naturally. For example, in a kitchen recording, a participant may be asked to prepare a simple pan-cooked dish by washing ingredients, cutting them on a board, heating a pan, spreading oil, manipulating food with a spatula or chopsticks, and transferring the result to a plate. The participant is not scripted at the frame level, but the task sequence ensures that the recording contains clear procedural boundaries, repeated hand--tool--object interactions, and observable state changes such as raw ingredients being cut, food being cooked, or tools being cleaned for later use. Such sessions are useful for benchmark construction because they provide temporally coherent clips with well-defined goals and visually grounded state transitions.

\paragraph{Open-ended participant-driven recording.}
In an open-ended participant-driven session, the participant is given only a broad activity goal and completes it in their own manner. For example, a participant may be asked to perform an unscripted household, craft, laboratory, or workshop activity using the tools they consider appropriate. This setting captures natural variation that is difficult to script, such as choosing a narrower tool because the target opening is small, replacing one tool with another after an unsuccessful attempt, or changing the manipulation strategy as the target object bends, tears, loosens, or becomes unstable. These recordings complement the structured sessions by exposing models to spontaneous tool choice, substitution, recovery, and state-dependent manipulation.

\subsection{Anonymized Contributor Statistics}
\label{app:contributor_statistics}

To preserve anonymity, we report aggregate role and background statistics for contributors from whom complete information was collected. These statistics cover a documented subset of the broader author and contributor pool and are not intended to enumerate every manuscript author or every individual acknowledged in Appendix~\ref{app:contributors}. We do not report names, contact information, institutional identifiers, public profiles, or exact per-video contribution counts.

\paragraph{Contributor roles.}
Contributors may participate in one or more roles, including video participant, narrator, QA annotator, QA reviewer, data curator, grounding annotator, or trajectory annotator. Since several contributors played multiple roles, the role counts in Table~\ref{tab:contributor_role_summary} are not mutually exclusive.

\begin{table}[!htbp]
\appendixtablestyle
\caption{\textbf{Contributor roles in the documented subset.} Counts are non-exclusive.}
\label{tab:contributor_role_summary}
\centering
\begin{tabularx}{\linewidth}{@{}l c X@{}}
\toprule
\textbf{Role} & \textbf{\# Contributors} & \textbf{Main Responsibility} \\
\midrule
Participant / narrator & 16 & Record egocentric videos and provide tool-centric narrations \\
QA annotator & 10 & Write human-crafted benchmark QA pairs \\
QA reviewer & 2 & Review answer validity, visual evidence, and distractor quality \\
Data curator & 3 & Curate videos, manage annotation progress, and organize benchmark data \\
Grounding annotator & 1 & Annotate focal-tool grounding for selected instances \\
Trajectory annotator & 2 & Construct or verify trajectory annotations for selected clips \\
\bottomrule
\end{tabularx}
\end{table}

\paragraph{Anonymization policy.}
These aggregate statistics are intended to document the range of contributor backgrounds and project roles without enabling linkage to real identities or specific released recordings. We do not release names, institutions, contact information, public profiles, or exact per-video contribution counts.

\begin{table}[!htbp]
\appendixtablestyle
\centering
\caption{
\textbf{Contributor backgrounds.}
Aggregate (a) degree and field and (b) age-group distributions for the documented subset.
}
\label{tab:contributor_backgrounds}

\begin{minipage}[t]{0.57\linewidth}
\vspace{0pt}
\centering
\textbf{(a) Degree and field}\\[0.5em]
\begin{tabularx}{\linewidth}{@{}X r X r@{}}
\toprule
\multicolumn{2}{c}{\textbf{Degree}} & \multicolumn{2}{c}{\textbf{Field}} \\
\cmidrule(lr){1-2}\cmidrule(lr){3-4}
\textbf{Category} & \textbf{\#} & \textbf{Category} & \textbf{\#} \\
\midrule
Undergrad. & 12 & Computer Science & 12 \\
Master's & 5 & Electrical Eng. & 7 \\
PhD & 5 & Biology & 2 \\
Postdoc & 1 & Chemistry & 1 \\
 & & Film / Media & 1 \\
\bottomrule
\end{tabularx}
\end{minipage}
\hfill
\begin{minipage}[t]{0.39\linewidth}
\vspace{0pt}
\centering
\textbf{(b) Age group}\\[0.5em]
\includegraphics[width=\linewidth]{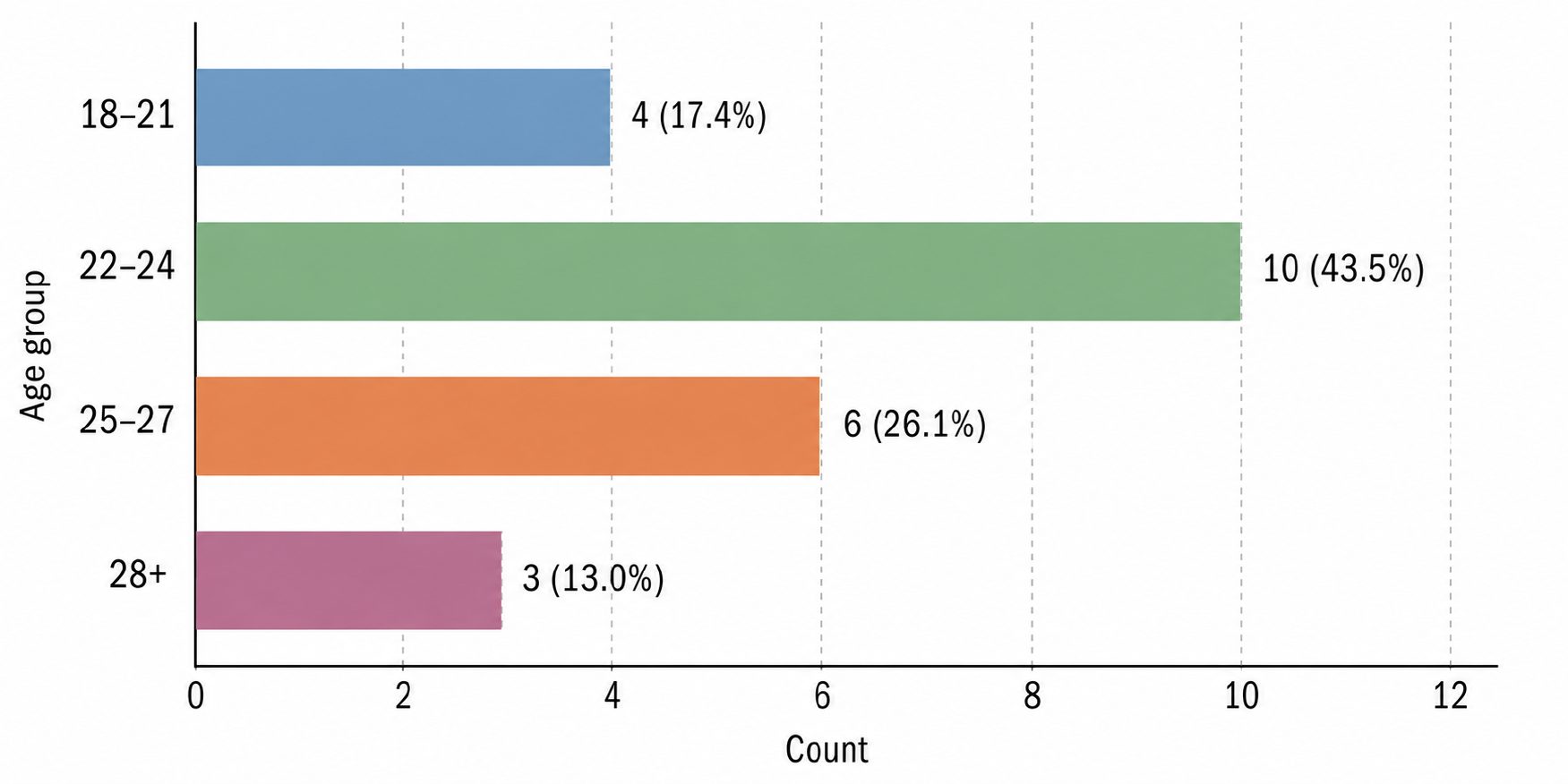}
\end{minipage}

\end{table}

\section{Preprocessing and Annotation Protocols}
\label{app:preprocessing_annotation}

\subsection{View Rectification Settings}
\label{app:view_rectification}

\paragraph{Recording device and modalities.}
EgoTools videos are recorded with HOMIE, a head-mounted multimodal recording platform that captures four synchronized fisheye camera streams together with audio and auxiliary motion and synchronization signals. These raw modalities provide the source data for downstream view rectification, temporal alignment, and selected geometry-related processing.

\paragraph{Canonical view construction.}
For annotation, instruction-data construction, benchmark construction, model evaluation, and public release, we use the rectified front-left RGB stream as the canonical egocentric view. The rectification process uses synchronized views from the same recording to produce a spatially normalized first-person representation while preserving the hand--tool--object interactions and surrounding workspace required for tool-use understanding.

\begin{figure}[!htbp]
    \centering
    \includegraphics[width=\linewidth]{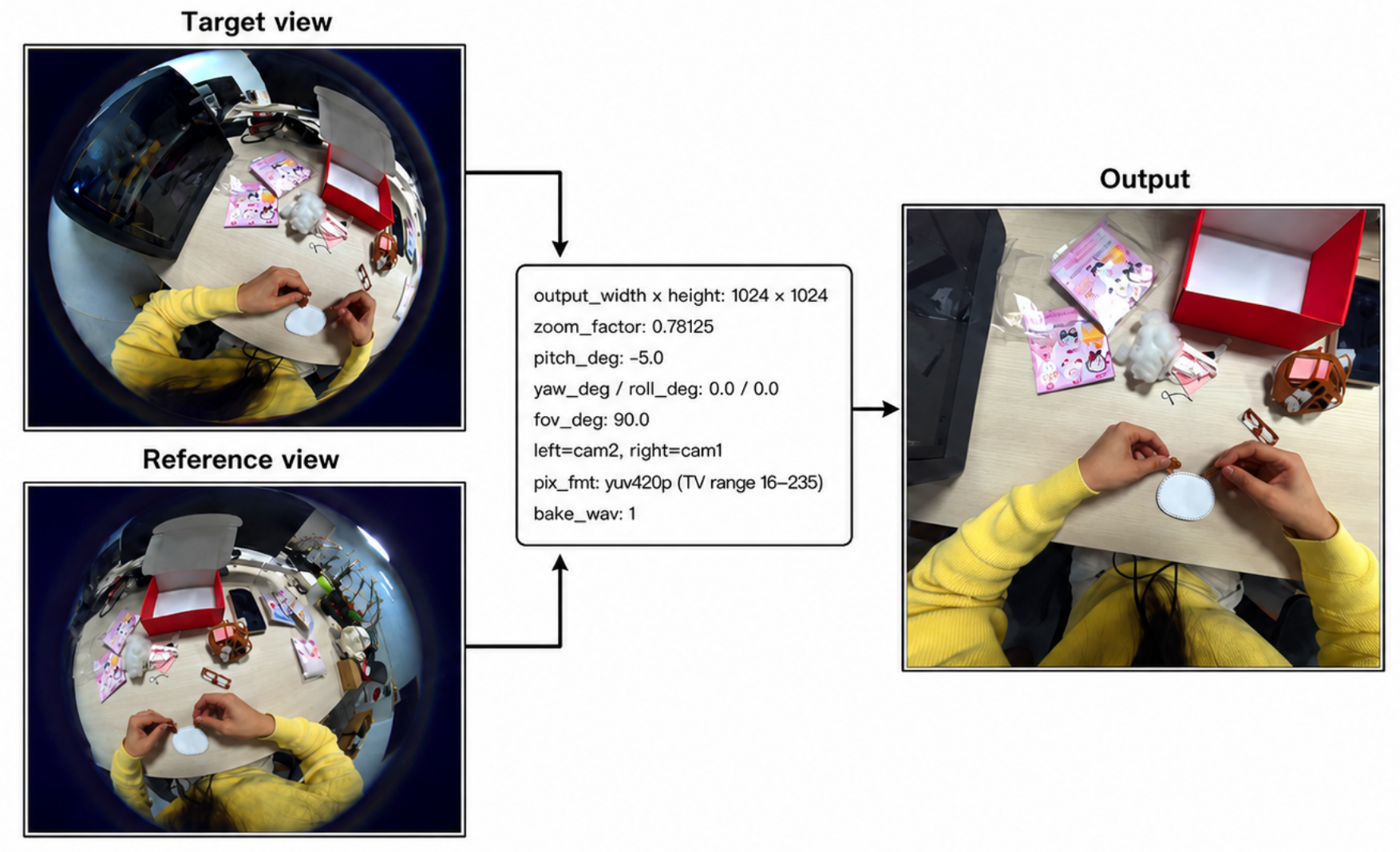}
    \caption{
    \textbf{Canonical-view rectification.}
    Example input views and the rectified first-person frame.
    }
    \label{fig:rectification}
\end{figure}

Figure~\ref{fig:rectification} shows a representative rectification example. The target and reference views are processed using fixed view parameters to produce the canonical output. For the displayed example, the output resolution is \(1024 \times 1024\), the zoom factor is \(0.78125\), the field of view is \(90.0^\circ\), and the pitch correction is \(-5.0^\circ\), with yaw and roll set to \(0.0^\circ\). The values shown in the figure correspond to the displayed example.

The resulting videos are single-view egocentric streams with a resolution of \(1024 \times 1024\) at 20 FPS and synchronized audio. The same canonical video representation is used for annotation, instruction-data construction, benchmark construction, model evaluation, and the publicly released video assets.

\paragraph{Release scope.}
The public release includes the rectified canonical videos rather than the original raw fisheye camera streams. We report a representative rectification configuration and the output specifications used in our data processing, while raw fisheye streams and internal preprocessing utilities are not included in the release.

\subsection{Narration Annotation Guidelines}
\label{app:narration_guidelines}

Participants provide narrations for meaningful tool-use events rather than densely describing every visible action. The goal is to capture which tool is being used, what target is being acted on, why the selected tool is appropriate in the current context, and what alternative tool may be relevant when applicable.

\paragraph{Annotation interface and procedure.}
The annotation interface allows participants or data collectors to watch rectified egocentric videos, identify relevant timestamps, and provide tool-centric narrations. Narrations are associated with temporally localized video segments and are written or corrected with reference to the visible hand--tool--object interaction.

The narration guidelines encourage annotators to specify the selected tool, a relevant alternative when applicable, the target object, the performed action, and the physical or procedural rationale for the tool choice. Narrations are free-form and need not follow a rigid sentence structure, but the following template is provided as a reference:

\begin{quote}
\small
\texttt{I am using [the] [property] [tool] [location/identity cue] rather than [the] [property] [alternative tool] [location/identity cue] to [action] on/for [target] because [reason].}
\end{quote}

\paragraph{Narration targets.}
Narrations focus on tool-use moments that involve meaningful decisions or state-dependent behavior, including selecting a tool, switching between tools, adapting manipulation to a changing object state, preparing a tool for a later step, recovering from an unsuccessful attempt, or using a tool in a physically appropriate manner. Participants are not required to narrate every visible action.

\paragraph{Narration language and normalization.}
Source narrations may be provided in English, Korean, or Chinese. Korean and Chinese narrations are transcribed and translated into English, and all narrations are subsequently normalized and corrected by bilingual or English-proficient project members when necessary. The corrected English text is paired with the corresponding video segment and used as the primary narration annotation signal for instruction-data construction.

Language tags retain information about the original narration language. Original narration audio is preserved as an internal dataset asset and may be included in the public release only when covered by the applicable release authorization and privacy review.

\paragraph{Benchmark language and evidence separation.}
All final EgoTools-Bench questions and answer choices are written in English. When benchmark construction relies on auxiliary narrations or rough temporal descriptions originally provided in Korean or Chinese, these materials are translated and normalized into English before QA construction or verification. Final benchmark items are reviewed for linguistic clarity, consistency with timestamped visual evidence, single-best-answer validity, and distractor quality.

Tool-centric narrations, corrected narration text, and narration audio are not provided to models during EgoTools-Bench evaluation. They are also not supplied to benchmark annotators as privileged answer evidence; benchmark questions must remain answerable from the corresponding video and question context.

\paragraph{Grounding-oriented annotations.}
Narrations may be associated with focal tools or target objects. For selected instances, annotators provide 2D grounding points on corresponding keyframes, linking textual mentions to visible object instances. These grounding annotations provide initialization cues for long-horizon tracking and support focal-tool localization, trajectory construction, and grounded video-language training.

Grounding annotations are available only for selected instances and are progressively released as annotation finalization, release authorization, and privacy review are completed.

\paragraph{Examples.}
\begin{itemize}
    \item I am using the narrow chopsticks rather than the spoon to reach the jam inside the jar because the spoon is too wide to fit through the opening.

    \item I am using the flat spatula rather than the chopsticks to lift the egg because the broad surface can support the soft egg without tearing it.

    \item I am using the pipette rather than the larger container to transfer a small amount of liquid because the pipette provides finer volume control.
\end{itemize}

\section{Dataset Statistics and Release}
\label{app:stats_release}

\subsection{Dataset and Benchmark Statistics}
\label{app:dataset_stats}

EgoTools contains 646 egocentric videos totaling 100.37 hours across seven tool-use domains: kitchen, classroom, research laboratory, repair workshop, craft, office, and household. Throughout the main paper, we refer to this corpus as approximately 100 hours. The corpus additionally contains 361,332 hierarchical captions and 6,519 tool-centric narrations. From these recordings, we reserve 40.34 hours of high-quality, tool-dense source videos exclusively for benchmark construction and evaluation.

\begin{table}[!htbp]
\appendixtablestyle
\centering
\caption{
\textbf{EgoTools dataset and benchmark statistics.}
}
\label{tab:supp_dataset_stats}
\begin{tabularx}{\linewidth}{@{}Xr@{}}
\toprule
\textbf{Statistic} & \textbf{Value} \\
\midrule
Total EgoTools videos & 646 \\
Total video duration & 100.37 hours \\
Hierarchical captions & 361,332 \\
Tool-centric narrations & 6,519 \\
Benchmark-reserved source duration & 40.34 hours \\
Benchmark QA pairs & 1,000 \\
Human-crafted QA pairs & 900 \\
Human-verified spatial QA pairs & 100 \\
\bottomrule
\end{tabularx}
\end{table}

\paragraph{Domain-level composition.}
Table~\ref{tab:domain_statistics} reports the distribution of the complete EgoTools-Data corpus and the corresponding domain allocation of EgoTools-Bench. Corpus statistics describe the full data collection, whereas benchmark QA counts describe the source domain assigned to each of the 1,000 benchmark questions.

\begin{table}[!htbp]
\appendixtablestyle
\centering
\caption{
\textbf{Domain composition of EgoTools-Data and EgoTools-Bench.}
The benchmark is not domain-balanced.
}
\label{tab:domain_statistics}
\begin{tabular*}{\linewidth}{@{\extracolsep{\fill}}lrrrrr@{}}
\toprule
& \multicolumn{4}{c}{\textbf{EgoTools-Data}}
& \multicolumn{1}{c}{\textbf{EgoTools-Bench}} \\
\cmidrule(lr){2-5}
\cmidrule(lr){6-6}
\textbf{Domain}
& \textbf{\# Videos}
& \textbf{Hours}
& \textbf{\# Captions}
& \textbf{\# Narrations}
& \textbf{\# QAs} \\
\midrule
Kitchen
& 276
& 32.37
& 116,532
& 2,113
& 837 \\

Classroom
& 74
& 6.86
& 24,696
& 467
& 15 \\

Research Lab
& 61
& 30.92
& 111,312
& 1,887
& 99 \\

Repair Workshop
& 84
& 12.34
& 44,424
& 763
& 16 \\

Craft
& 67
& 10.17
& 36,612
& 657
& 23 \\

Office
& 30
& 2.83
& 10,188
& 213
& 6 \\

Household
& 54
& 4.88
& 17,568
& 419
& 4 \\
\midrule
\textbf{Total}
& \textbf{646}
& \textbf{100.37}
& \textbf{361,332}
& \textbf{6,519}
& \textbf{1,000} \\
\bottomrule
\end{tabular*}
\end{table}

The domain distribution highlights the distinction between corpus coverage and benchmark composition. EgoTools-Data contains substantial recordings from all seven domains, including approximately 31 hours of research-laboratory activity and more than 12 hours of repair-workshop activity. EgoTools-Bench, however, is dominated by Kitchen questions and should not be interpreted as a domain-balanced evaluation set. The benchmark was constructed primarily to cover complementary tool-use reasoning capabilities, annotation quality, and challenging tool-mediated interactions rather than to support statistically balanced comparisons across collection domains. Domain-level results for the smallest benchmark slices would therefore be unstable and are not used for primary performance claims.

\paragraph{Reasoning-track distribution.}
EgoTools-Bench contains \textbf{1,000} 8-way multiple-choice question--answer pairs, including \textbf{900} human-crafted questions and \textbf{100} human-verified spatial questions generated from the 3D annotation pipeline. Across the four research-facing tracks, the benchmark includes 363 Affordance \& Causality questions, 236 Perception \& Grounding questions, 222 Procedural Dynamics questions, and 179 Spatial Reasoning questions. This distribution reflects the benchmark's emphasis on tool selection, application, and physical effects while maintaining coverage of perceptual grounding, procedural understanding, and spatial hand--tool--object relations.

\paragraph{Evaluation-only benchmark partition.}
All benchmark source videos, clips, questions, answers, and derived annotations are reserved exclusively for evaluation. No clip, caption, narration, synthetic QA, or other annotation derived from a benchmark-reserved source video is included in the instruction-tuning corpus.

\subsection{Release and Privacy Considerations}
\label{app:release_privacy}

The EgoTools release is scoped to research use and includes only materials covered by the applicable participant consent, release authorization, anonymization procedures, and privacy review. The release provides processed rectified egocentric videos rather than the original raw fisheye recordings, together with the annotations and metadata required for benchmark evaluation and dataset documentation. Additional consent and privacy safeguards are described in Appendix~\ref{app:ethics}.

\paragraph{Initial release components.}
The initial release package will include the following cleared components:

\begin{itemize}[leftmargin=*]
    \item Rectified egocentric video clips;
    \item Dense hierarchical captions;
    \item Corrected English tool-centric narrations;
    \item All 1,000 EgoTools-Bench question--answer pairs and their evaluation annotations;
    \item Clip metadata and timestamps;
    \item Source-video-disjoint training and benchmark split information; and
    \item Dataset documentation and benchmark evaluation metadata.
\end{itemize}

\paragraph{Incrementally released components.}
The following components are released incrementally as their annotation, authorization, and privacy-review procedures are completed:

\begin{itemize}[leftmargin=*]
    \item Original narration audio for consent-cleared examples;
    \item Selected finalized focal-tool grounding annotations;
    \item Selected object-state and trajectory annotations; and
    \item 3D trajectory visualizations for selected clips.
\end{itemize}

The availability of these components may differ across examples because grounding, audio, and trajectory assets require additional annotation-finalization and release-review procedures.

\paragraph{Excluded components.}
Raw fisheye camera streams are not publicly released. The release also excludes personally identifying metadata, uncleared recordings or audio, sensitive or accidentally captured content, internal identity mappings, exact per-video contributor counts, and internal preprocessing utilities.

\paragraph{Privacy safeguards.}
Names, contact information, institution-specific identifiers, public-profile information, and other personally identifying information are removed or withheld from released metadata. Sensitive, private, or accidental content is reviewed before release, and materials deemed unsuitable for public distribution are excluded. Contributors may request exclusion of data associated with their participation in accordance with the release policy and applicable consent terms.

\paragraph{Access conditions and license.}
EgoTools is distributed under a research-use license specifying permitted uses, redistribution conditions, and usage restrictions. Privacy-sensitive or conditionally released components are handled according to their applicable authorization and privacy-review outcomes. The dataset is not intended for participant identification, biometric profiling, surveillance, or other uses inconsistent with the original collection and release scope.

\paragraph{Limitations.}
Although EgoTools-Data spans seven real-world tool-use domains, it does not exhaustively cover all tools, environments, participant populations, cultural practices, or professional procedures. The benchmark distribution is strongly concentrated in Kitchen videos and should not be treated as representative of the domain distribution of the full corpus. Grounding, narration-audio, and 3D trajectory components are available only for examples that have passed the corresponding annotation-finalization, release-authorization, and privacy-review procedures.

\section{Quality Control and Normalization}
\label{app:qc}

\subsection{Fix-First Quality Control}
\label{app:qc_fix_first}

Raw human annotations are processed under a fix-first quality-control policy that prioritizes repairing valid annotations before discarding them. Items are retained whenever deterministic or minimal model-assisted edits can preserve the original annotator intent without introducing ambiguity.

\paragraph{Structural audit.}
Each item is evaluated using a 22-flag binary audit covering defects such as placeholders, missing fields, malformed options, duplicate entries, first-person leakage, personally identifiable information, answer--option mismatch, inappropriate full-video clip usage, and distractor anomalies.

\paragraph{Filtering policy.}
Content-empty submissions and verbatim duplicates are removed. Repairable defects are corrected through deterministic rules or minimal model-assisted edits. Cases for which the intended meaning or unique answer cannot be established with high confidence are routed to human review rather than being automatically accepted.

\paragraph{Model-assisted repair.}
We use Gemini-2.5-Flash-Lite with temperature 0 for model-assisted repair. The repair prompt instructs the model to perform the smallest meaning-preserving edit possible. Typical edits include replacing personally identifying names with role-based references, rewriting first-person expressions into third-person form, correcting surface-level grammar, and completing truncated answer text only when the intended content is unambiguous. Edits that could alter the semantic content or answer validity are routed to human review.

\subsection{Eight-Choice Normalization and Anti-Shortcut Checks}
\label{app:qc_normalization}

\paragraph{Eight-choice normalization.}
All benchmark items are normalized to exactly eight answer options. We dispatch each item according to its current distractor count: items with fewer than seven distractors are augmented, items with exactly seven distractors are validated, and items with more than seven distractors are reduced to seven plausible alternatives. This process produces a consistent 8-way multiple-choice format while preserving the original correct answer and question intent whenever possible.

\paragraph{Anti-shortcut checks.}
Each normalized item and the resulting benchmark corpus are subjected to deterministic anti-shortcut checks before finalization. Item-level checks detect meta-options such as ``all of the above'' or ``none of the above,'' duplicate or near-duplicate options, token-set permutations of the correct answer, substring or superstring containment, and excessive question--answer lexical overlap. Corpus-level checks additionally examine answer-position imbalance and systematic answer-length bias. Failed items are routed to bounded regeneration or human review, with detected violations provided as feedback for revising the answer options.

\paragraph{Length-bias check.}
For each item, we compare the character length of the correct answer with the distribution of distractor lengths. Let \(L_{\mathrm{ans}}\) denote the character length of the correct answer, and let \(\mu_{\mathrm{dist}}\) and \(\sigma_{\mathrm{dist}}\) denote the mean and standard deviation of the seven distractor lengths. We compute
\[
z =
\frac{L_{\mathrm{ans}}-\mu_{\mathrm{dist}}}
{\sigma_{\mathrm{dist}}}.
\]
Items satisfying \(\lvert z\rvert>1.5\) are flagged for repair. Previously cached eight-option snapshots are subjected to the same checks rather than being accepted without revalidation, ensuring that the length constraint is consistently enforced across all benchmark items. An earlier cached snapshot exhibited a corpus-level mean standardized deviation of \(\bar{z}\approx+2.21\). After applying the finalized check and repair procedure, the retained benchmark has \(\bar{z}\approx-0.15\), with every item satisfying \(\lvert z\rvert\leq1.5\).

\paragraph{Option shuffling.}
The final answer-option order is shuffled using a deterministic seed derived from each QA identifier. This reduces positional priors while keeping evaluation reproducible. Because every released benchmark item contains eight answer choices, the main benchmark tables report standard multiple-choice accuracy with a consistent chance baseline of 12.5\%.

\FloatBarrier

\section{Additional Related Work}
\label{app:additional_related_work}

\paragraph{Broader egocentric video reasoning.}
Egocentric video benchmarks have expanded from action and object recognition toward episodic memory, planning, temporal localization, situated reasoning, and assistance. QAEgo4D and GroundVQA study episodic-memory QA and temporally grounded QA in long egocentric videos~\citep{barmann2022qaego4d,di2024groundvqa}. EgoTaskQA evaluates task-oriented questions about dependencies, effects, goals, and beliefs~\citep{jia2022egotaskqa}; EgoSchema and EgoTempo emphasize long-context and temporal reasoning~\citep{mangalam2023egoschema,plizzari2025egotempo}; and EgoPlan-Bench evaluates planning from first-person videos~\citep{chen2024egoplanbench}. More recent benchmarks examine first-person thinking, life-log and real-time assistance, expert--trainee support, situated awareness, multi-agent egocentric QA, intent understanding, personalized grounding, and AR or life-logging environments~\citep{cheng2024egothink,gao2026lifeeval,kim2026maegoqa,li2026situated,pan2026egointent,ragusa2025egoextra,tang2026egoeverything,xiao2026egogrounding,yang2025egolifeegocentriclifeassistant,zhou2025xlebench}. These resources substantially broaden egocentric evaluation around memory, temporal context, planning, and user assistance. However, they generally do not isolate the physical reasoning required to determine why a particular tool is appropriate, which substitute is feasible, or how manipulation should adapt as the target object changes state.

\paragraph{Instructional, embodied, and robotic interaction benchmarks.}
Instructional-video datasets provide useful comparisons because they frequently contain tools, ingredients, techniques, and alternative procedures. VidDetours retrieves procedural detours from how-to videos, while StepDiff identifies differences between instructional clips~\citep{ashutosh2024detours,nagarajan2024stepdiff}. Their primary focus, however, is procedural retrieval or comparison rather than physical tool-use reasoning grounded in a continuous first-person task. Embodied interaction benchmarks provide another related but distinct setting. OpenEQA evaluates open-vocabulary embodied question answering in real-world environments~\citep{majumdar2024openeqa}, RoboVQA provides large-scale video-language supervision from robotic demonstrations~\citep{sermanet2023robovqa}, and RoboCasa365 studies simulated kitchen manipulation across diverse tasks and environments~\citep{nasiriany2026robocasa365}. These resources offer important scope contrasts, but they do not provide the same combination of natural human egocentric video, tool-centric narration, narration-linked focal-tool grounding, and explicit QA over tool choice, substitution, ordered manipulation, and state-dependent adaptation.

\paragraph{Affordance grounding and physical tool understanding.}
Robotics and embodied AI have long treated tool use as a problem of affordance, grasping, motion, and task execution. Task-oriented grasping studies how tools should be grasped for activities such as sweeping and hammering~\citep{fang2019taskoriented}, while ToolFlowNet predicts dense tool motion for manipulation tasks such as scooping and pouring~\citep{seita2022toolflownet}. Broader surveys and manipulation methods study robot tool use, affordance-based manipulation, precise affordance grounding, and physically grounded policies~\citep{brohan2023rt2,gao2024physically,li2025learningprecise,qin2023robottooluse,yamanobe2017brief}. Recent multimodal benchmarks and methods further evaluate whether MLLMs and VLMs can recognize affordances, infer physical constraints, and ground tool-related interactions~\citep{huang2024manipvqa,qian2024affordancellm,yu2025seqafford,zhang2025phystoolbench,yao2026hammer}. These studies reveal persistent weaknesses in reasoning about tool functions, physical constraints, and multi-step interactions. Nevertheless, many evaluations are based on static images, synthetic or reconstructed 3D scenes, short interactions, or constrained robotic settings. EgoTools-Bench is complementary: it evaluates physical tool-use reasoning from natural first-person videos in which models must interpret real human tool selection, substitution, manipulation, and adaptation over time.

\paragraph{Wearable assistants.}
Wearable and egocentric assistant benchmarks are especially relevant because they require models to answer situated questions from the user's viewpoint. WearVQA studies visual question answering for wearable devices~\citep{chang2025wearvqa}, while visual-intention grounding evaluates whether models can infer user needs and intentions from egocentric observations~\citep{sun2025visualintention}. These settings motivate first-person assistance but do not specifically center tool-mediated physical reasoning. EgoTools-Bench therefore complements wearable-assistant research by focusing on questions that require understanding the functional role of tools, feasible substitutes, temporally ordered manipulation, and adaptation to changing object states.

\section{Training and Evaluation Details}
\label{app:training_details}

\subsection{Single-Stage Training Procedure}
\label{app:two_stage_training}

\paragraph{Training mixture.}
We use the 184,679-example corpus described in Section~\ref{ssec:sft_data}. By answer format, the mixture contains 170,595 multiple-choice examples, 5,000 short open-ended examples, and 9,084 dense-captioning and narration-completion examples; answer letters in the next-action subset are balanced across the four positions at 1,250 examples each. Table~\ref{tab:training_data_composition} gives the capability-level breakdown.

The next-action and single-image open-ended examples are generated from EgoTools training videos and their caption timelines; no video, image, or QA annotation from EgoSchema, EgoPlan-Bench, or EgoThink is used. A subset of their question strings coincides with generic benchmark question templates, while media overlap with the benchmark reference set is zero.

\paragraph{Optimization.}
We perform full-parameter supervised fine-tuning of the language-model component of Qwen3-VL-8B-Instruct, while keeping the vision encoder and multimodal projector frozen. Thus, ``Full SFT'' refers to full-parameter updating of the language-model component rather than updating every module in the multimodal architecture.

We optimize using AdamW with a learning rate of \(2.3\times10^{-6}\), a constant learning-rate schedule, and no warmup. Training is performed for one epoch with a per-device batch size of 1, gradient accumulation over 16 steps, and 8 GPUs, yielding an effective batch size of 128 and 1,443 optimizer steps. We use bfloat16 precision, DeepSpeed ZeRO Stage 3, FlashAttention, and gradient checkpointing.

Each training video is represented by 64 uniformly sampled frames at a target rate of 2 FPS, with the per-video frame count fixed to 64. We cap each image at 1,024 visual tokens and use a maximum multimodal sequence length of 8,192 tokens.

\paragraph{Source-video separation.}
The EgoTools training pool and EgoTools-Bench are partitioned before instruction-data or benchmark construction. No video, clip, caption, narration, synthetic QA pair, grounding annotation, or other derivative of a benchmark-reserved source video is used during training.

\par\medskip
\noindent
\begin{minipage}{\linewidth}
\subsection{Training Data Composition}
\label{app:training_data_composition}

The training procedure uses 184,679 EgoTools-derived examples. Table~\ref{tab:training_data_composition} summarizes their composition by the capability each example supervises.

\par\medskip
\captionsetup{type=table}
\appendixtablestyle
\centering
\caption{
\textbf{Training data by supervised capability.}
Shares of the 184,679-example mixture are rounded to one decimal place.
}
\label{tab:training_data_composition}
\begin{tabularx}{\linewidth}{@{}Xrr@{}}
\toprule
\textbf{Supervised capability}
& \textbf{\# Examples}
& \textbf{Share} \\
\midrule
Affordance \& Causality (AC):
tool-use purpose, intent, and causal rationale
& 14,175
& 7.7\% \\

Perception \& Grounding (PG):
tool and state identification, attributes, and frame-level visual evidence
& 14,755
& 8.0\% \\

Procedural Dynamics (PD):
temporal order, workflow, and state change within a clip
& 22,943
& 12.4\% \\

Spatial Reasoning (SR):
spatial relations, placement, and hand--tool--object geometry
& 17,887
& 9.7\% \\
\midrule
Episode-level multiple choice:
action ordering, activity summarization, and goal inference over a whole
episode (64,554), plus next-action questions paired with the current
observation frame (5,000)
& 69,554
& 37.7\% \\

General video QA:
mixed episode-level questions not targeted at a single ability
& 31,281
& 16.9\% \\

Dense captioning and narration completion
& 9,084
& 4.9\% \\

Single-image open-ended QA
& 5,000
& 2.7\% \\
\midrule
\textbf{Total}
& \textbf{184,679}
& \textbf{100.0\%} \\
\bottomrule
\end{tabularx}
\end{minipage}
\par\medskip

\subsection{Evaluation Protocols}
\label{app:checkpoint_evaluation}

The results in Table~\ref{tab:sft_results} use the standard 64-frame MP4-based protocol for EgoTools-Bench. EgoSchema is evaluated on the full 5,031-question set, EgoPlan-Bench on its 3,343-question validation set, and EgoThink using macro-averaged accuracy over its twelve single-image subtasks, judged by \texttt{gpt-4o-2024-11-20}.

\end{document}